\documentclass[10pt,journal,compsoc]{IEEEtran}
\usepackage{amsmath,amsfonts}
\usepackage{algorithmic}
\usepackage{array}
\usepackage{cite}
\usepackage{textcomp}
\usepackage{stfloats}
\usepackage{url}
\usepackage{verbatim}
\usepackage{graphicx}
\usepackage{amsmath}
\usepackage{nccmath}
\usepackage{mathtools}
\usepackage{amssymb}
\usepackage{xcolor}
\newcommand{\bb}[1]{\boldsymbol{#1}}
\DeclareMathOperator*{\argmax}{arg\,max}

\usepackage{algorithm}
\usepackage{algorithmic}
\usepackage{tabularx}
\usepackage{graphicx}
\usepackage{comment}
\usepackage{caption}
\usepackage{subcaption}
\usepackage[justification=centering]{caption}
\def\BibTeX{{\rm B\kern-.05em{\sc i\kern-.025em b}\kern-.08em
    T\kern-.1667em\lower.7ex\hbox{E}\kern-.125emX}}
\usepackage{balance}

\begin{document}

\title{FiMReS$t$: Finite Mixture of Multivariate Regulated Skew-$t$ Kernels - A Flexible Probabilistic Model for Multi-Clustered Data with Asymmetrically- Scattered Non-Gaussian Kernels}

\author{Sarmad~Mehrdad,~\IEEEmembership{Student Member,~IEEE,}
        S. Farokh~Atashzar,~\IEEEmembership{Senior Member,~IEEE}% <-this % stops a space
\IEEEcompsocitemizethanks{\IEEEcompsocthanksitem Sarmad Mehrdad is with the Department of Electrical and Computer Engineering, Tandon School of Engineering, New York University, New York, NY, 11201.\\
S. Farokh Atashzar is with the Department of Mechanical and Aerospace
Engineering and the Department of Electrical and Computer Engineering, and
the Department of Biomedical Engineering, Tandon School of Engineering, New York University, New York, NY, 11201. He is also with NYU WIRELESS and NYU
CUSP.\protect\\
E-mail: sfa7@nyu.edu%
}}

% The paper headers
% \markboth{IEEE TRANSACTIONS ON PATTERN ANALYSIS AND MACHINE INTELLIGENCE}%
% {Mehrdad \MakeLowercase{\textit{et al.}}: FiMReS$t$: Finite Mixture of Multivariate Regulated Skew-$t$ Kernels - A Flexible Probabilistic Model for Multi-Clustered Data with Asymmetrically- Scattered Non-Gaussian Kernels}

\IEEEtitleabstractindextext{%
\begin{abstract}
Recently skew-$t$ mixture models have been introduced as a flexible probabilistic modeling technique taking into account both skewness in data clusters and the statistical degree of freedom (S-DoF) to improve modeling generalizability, and robustness to heavy tails and skewness. In this paper, we show that the state-of-the-art skew-$t$ mixture models fundamentally suffer from a hidden phenomenon named here as “S-DoF explosion,” which results in local minima in the shapes of normal kernels during the non-convex iterative process of expectation maximization. For the first time, this paper provides insights into the instability of the S-DoF, which can result in the divergence of the kernels from the mixture of $t$-distribution, losing generalizability and power for modeling the outliers. Thus, in this paper,  we propose a regularized iterative optimization process to train the mixture model, enhancing the generalizability and resiliency of the technique. The resulting mixture model is named Finite Mixture of Multivariate Regulated Skew-$t$ (FiMReS$t$) Kernels, which stabilizes the S-DoF profile during optimization process of learning. To validate the performance, we have conducted a comprehensive experiment on several real-world datasets and a synthetic dataset. The results highlight (a) superior performance of the FiMReS$t$, (b) generalizability in the presence of outliers, and (c) convergence of S-DoF.
\end{abstract}

}

\maketitle

\IEEEdisplaynontitleabstractindextext

\IEEEpeerreviewmaketitle

\IEEEraisesectionheading{\section{Introduction}\label{intro}}

\IEEEPARstart{C}{lustering} algorithms have been utilized broadly in the literature for unsupervised learning, decoding the inherent clusters of a given data or signal space to better understand underlying hidden nature and patterns and to also detect anomalies, outliers, and specific abnormal trends \cite{61, 62, 63, 65}. The applications can range from fault detection in industries \cite{4,52,53}, up to cancer detection in health sectors \cite{2,3}, and even recently in robot learning where clustering methods are used to model the most probable behavior of a human trainer, in the context of learning from demonstration \cite{1,40, 58, 59}.
Classic clustering methods range from simple algorithms such as K-means \cite{6} up to more comprehensive methods using different probabilistic kernels such as Gaussian Mixture Models (GMM) \cite{7}. Probabilistic clustering using GMM was first suggested for speaker identification \cite{7,8,9} and has gained significant attention from numerous researchers for data segregation in applications such as biological data evaluations \cite{10}, robotics \cite{11,41}, and pattern recognition \cite{12,42}. 

The prominent iterative algorithm that is used to train a GMM is Expectation-Maximization (EM) approach \cite{13} that aims to maximize the model’s fitness (i.e., log-likelihood) at each iteration until convergence. However, it is imperative to note that elliptical distribution, such as Gaussians used in GMMs, will impose symmetricity assumption to the kernels of the probabilistic mixture model, challenging the performance of the model in the presence of skewness, heavy tails, and outliers, all of which are common in real-world datasets \cite{26}.

Thus, there has been an accelerated surge in proposing non-normal mixture modeling approaches such as mixture models based on generalized hyperbolic distribution \cite{23}, general split Gaussian distribution \cite{24}, and generalized hyperbolic factor analyzers \cite{64}. Details can be found in \cite{25} and references therein.  Lin et al. proposed a  solution by designing the skew-normal mixture model (SNMM) \cite{14} based on multivariate skew normal (MSN) distribution and the corresponding explicit EM algorithm, which mathematically relaxes the dependency of mixture modeling on the elliptical nature of Gaussian kernels.

\begin{figure}[h]
    \centering\includegraphics[width=\columnwidth]{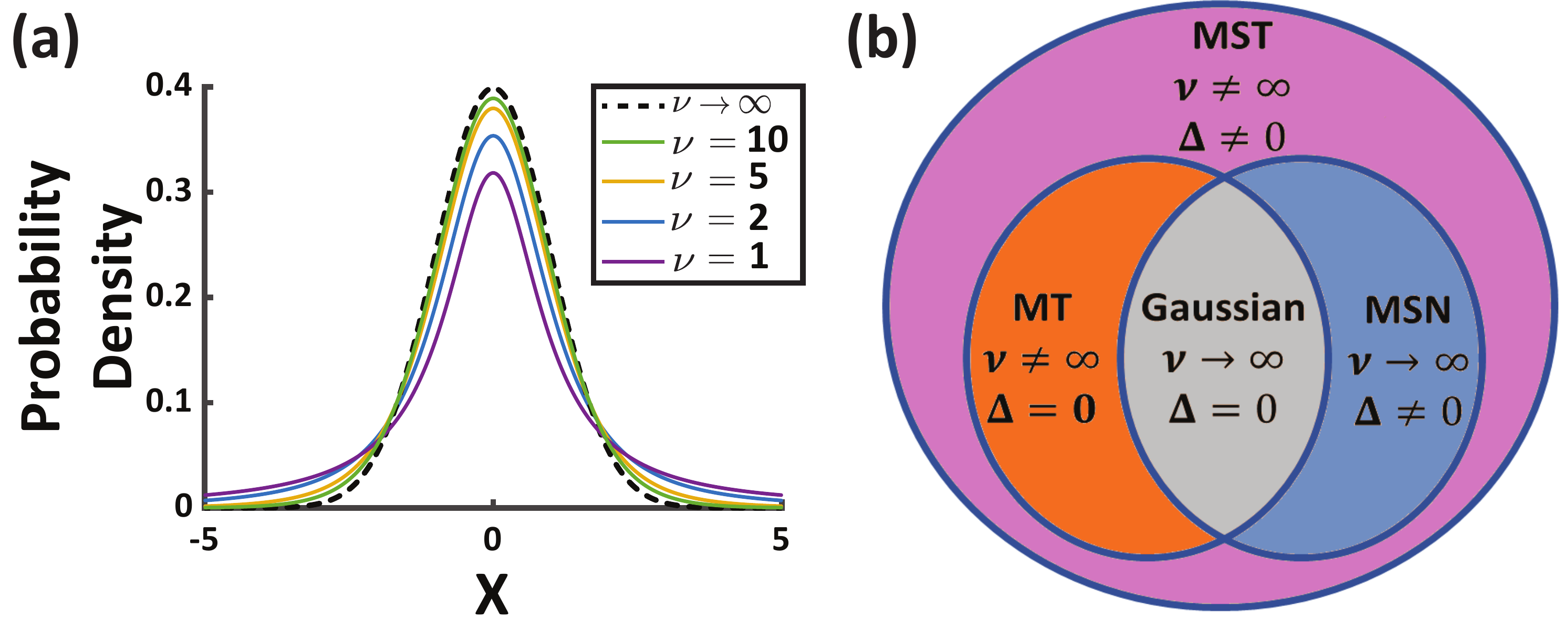}
    \caption{(a) Comparison of $t$-distribusions with zero mean, unit variance, and various S-DoF values. When $\nu \to \infty$, the probability density will become a Gaussian distribution. (b) Venn diagram visualizes the relationship between multivariate skew-$t$ distribution over MT, MSN, and Gaussian distributions. $\bb{\Delta}$ indicates the skewness in distribution.}
    \label{Fig1_new}
\end{figure}

In order to further improve the SNMM technique and to accommodate for the datasets with outliers and heavy tails, noted in the literature \cite{26}, a series of efforts was initiated \cite{16} to fundamentally replace the multivariate skew normal kernel with the multivariate skew-$t$ (MST) distribution kernel, which is a skewed version of the multivariate $t$ (MT) distribution. This attempt is due to the higher “flexibility in reach” and generalizability that the skew-$t$ distribution enables by incorporating the statistical degree of freedom (S-DoF) denoted in the literature by $\nu$.  Fig. \ref{Fig1_new} shows the schematic view of the generalizability of the multivariate skew-$t$ over multivariate skew normal, multivariate $t$, and Gaussian distributions. Thus, ideally, the mixture modeling techniques derived from skew-$t$ distribution would be more reliable for modeling a mixture of clusters subject to skewness, outliers, heavy tails, and data dropout \cite{54}.

The EM algorithm for training a mixture of multivariate skew-$t$ distributions cannot be implemented in closed-form since some conditional expected values involved in this approach are mathematically intractable. Thus, there were several attempts (such as \cite{26} and \cite{51}) to train a mixture model based on a simplified formulation of skew-$t$ distribution, which reduces the dimensionality of the skewness matrix down to unity. Such simplification allowed researchers to propose an explicit solution \cite{21,18,20} and thus realize an estimate of skew-$t$ mixture modeling. However, as expected, the aforementioned simplification limited the ability and flexibility of the resulting EM algorithm in modeling the skewness of the kernels. 

In order to expand the previous simplified versions of the mixture models based on skew-$t$ distribution to harness its inherent flexibility, in \cite{15}, Lin proposed the use of the Monte Carlo EM (MCEM) algorithm \cite{43} to numerically estimate the expected values needed in the corresponding EM approach. However, the resulting method suffered from slow convergence and inaccurate expectations due to the numerical nature of MCEM. %In order to improve the solution mentioned above, Ho et al. introduced an EM algorithm in \cite{48} by expressing two (out of three) of the intractable expected values in terms of moments of multivariate truncated $t$-distributions.
Finally, in an extended effort, Lee et al. \cite{19} proposed a closed-form EM algorithm for training skew-t mixture model (STMM), using a new formulation for estimating one of the intractable conditional expected values of the latent variables using one-step-late (OSL) estimation approach \cite{49}, while using the formulations proposed by Ho et al. \cite{48}, to express the other two intractable expected values (using the first and second moments of truncated multivariate $t$-distribution).

However, in \cite{19}, the skewness matrix was assumed to be diagonal. This restriction was solved in \cite{22} by proposing a finite mixture of canonical fundamental skew-t (FM-CFUST) distribution using the OSL-based EM algorithm to incorporate full skewness matrices for the clusters. This model was designed aiming for the highest optimality compared to its counterparts since it exploits the complete skewness matrix as opposed to the STMM model. 

Although probabilistic clustering has shown to be effective in modeling multivariate datasets, the numerical methods utilized for fitting such probabilistic models can be suboptimal. In this paper, for the first time, we will show that due to the non-convex nature of the existing iterative optimization approaches (such as the one in \cite{22})  for skew-$t$ kernels, there is a hidden phenomenon called here as \emph{S-DoF Explosion}. This condition is caused by the fact that any $t$-distribution can converge to a normal distribution when the S-DoF goes to infinity. This feature makes the non-convex iterative optimization prone to local minima caused by suboptimal normal or skew-normal kernels. No mixture modeling based on $t$-distribution exists that concurrently can escape local minima caused by the S-DoF explosion while accounting for non-diagonal and heavy skewness, non-diagonal covariance, high dimension, and dense data.

Thus, in this paper, a new mathematical formulation is developed for an iterative optimization that can generate a mixture model of non-diagonally-skewed kernels with a ``regulated S-DoF'' which addresses (a) restrictive assumptions on data dimension, (b) restrictive assumptions on covariance diagonality, (c) the neglection of heavy skewness, and (d) the assumption of diagonal skewness. 

In this paper, to address these issues, an EM-type algorithm called expectation-regularization-maximization (ERM) is developed. Subsequently, a finite mixture of regularized skew-t kernels (FiMReS$t$) distribution is generated, which can model a mixture of skew-$t$ distribution while avoiding S-DoF explosion. The algorithm was tested on asymmetrically scattered datasets to validate the performance. The results show that the proposed approach has a superior behavior taking advantage of a bounded S-DoF with minimum to no sensitivity to initialization as opposed to the counterparts in the literature. %The comparative results of fitting this mixture model over various datasets using this approach show that FiMReS$t$ has a superior performance.  

The rest of this paper is organized as follows. In Section II, the preliminaries are provided. In Section III, we mathematically design the proposed method modifying the maximum likelihood estimation (MLE) process. In section IV, the experiments evaluating the performance of the proposed method are given and comparative investigations are conducted.

\section{Preliminaries}\label{prel}

In this paper, random variables are shown by upper case letters such as $X$, and the corresponding values are shown in lower case. Also, where $z$ indicates a scalar value, boldened letters $\bb{z}$ are used to indicate vectors or matrices. 

In the literature, a weighted summation of multiple multivariate skew-$t$ distributions is shown as seen in Eq.~\eqref{CFUST} \cite{22}.

\begin{equation}
    \label{CFUST}
    f(\bb{y|\Theta}) = \sum^{g}_{i = 1}\omega_iMST_{p,\nu_i}(\bb{y}|\bb{\mu}_i,\bb{\Sigma}_i,\bb{\Delta}_i)
\end{equation}
\noindent
subject to
\begin{equation}
    \label{(4)}
    \sum^{g}_{i = 1}\omega_i = 1
\end{equation}

\noindent
In Eq.~\eqref{CFUST}, $g$ is the number of $p$-dimensional multivariate skew-$t$ distribution clusters in the mixture model, denoted as $MST_{p,\nu}(\boldsymbol{\mu, \Sigma, \Delta})$, and $i$ subscript addresses the parameters associated with the $i^{th}$ cluster in the mixture model. In addition, $\bb{y}$ is the input vector, and $\bb{\Theta}$ is the set that contains the defining parameters for all multivariate skew-$t$ kernels in the mixture model. The defining parameters for each cluster include  a $p \times 1$ mean vector $\bb{\mu}$, a $p \times p$ symmetric positive-definite covariance matrix $\bb{\Sigma}$, a $p \times p$ skewness matrix $\bb{\Delta}$, and a scalar value $\nu$ as S-DoF.

The multivariate skew-$t$ distribution, denoted as $MST_{p,\nu}(\boldsymbol{\mu, \Sigma, \Delta})$ in Eq.~\eqref{CFUST}, is a member of the skew elliptical distribution family \cite{26}. The PDF of this distribution is calculated using Eq. \eqref{(1)} as can be seen in

\begin{multline}
    \label{(1)}
    MST_{p,\nu}(\bb{\mu, \Sigma, \Delta}) = 2^p t_{p,\nu}(\bb{y|\mu,\Omega})T_{p,\nu'}(\bb{y^{\star}|0,\Lambda}).
\end{multline}

In Eq.\eqref{(1)}, $t_{p,\nu}(\bb{y|\mu,\Omega})$ is the PDF of the $p$-dimensional $t$-distribution ($t$-PDF) for random vector $\bb{y}$, where $\bb{\mu}$ is a $p\times 1$ mean vector, $\bb{\Omega}$ is a positive-definite symmetrical $p\times p$ covariance matrix, and $\nu$ is the scalar S-DoF parameter which can range from 2 to infinity for multivariate cases. In addition, $T_{p,\nu'}(\bb{y^{\star}|0,\Lambda})$ is the $p$-dimensional $t$-distribution cumulative density function ($t$-CDF) for a random vector $\bb{y^{\star}}$, given $\bb{0}$ as the mean, $\bb{\Lambda}$ as positive-definite symmetrical covariance matrix, and $\nu'$ as the S-DoF.

This distribution represents a general class from which different PDFs can be obtained. When there is zero skewness (i.e., $\bb{\Delta = 0}$), the multivariate skew-$t$ PDF represents a multivariate $t$-distribution, and when S-DoF becomes a large scalar (i.e., $\nu \rightarrow \infty$), the multivariate skew-$t$ PDF converge to the multivariate skew-normal distribution. In the case where there is no skewness and the S-DoF is large, the multivariate skew-$t$ distribution will become a Gaussian distribution $N_p(\bb{x}|\bb{\mu},\bb{\Sigma})$. (Seen in Fig.~\ref{Fig1_new}(b)).

\noindent
 The parameters ($\bb{\Omega}$, $\bb{y^{\star}} $, $\bb{c(y)} $, $\bb{\Lambda}$, $\bb{d(y})$, $\nu'$) in Eq.~\eqref{(1)} defines the multivariate skew-$t$ PDF and are explained as follows. 

\begin{align*}
    &\bb{\Omega} = \bb{\Sigma + \Delta^T\Delta}\\
    &\bb{y^{\star}} = \bb{c(y)}\sqrt{\frac{\nu + p}{\nu + d(\bb{y})}}\\
    &\bb{c(y)} =\bb{\Delta^T\Omega^{-1}(y - \mu)}\\
    &\bb{\Lambda} = \bb{I_p - \Delta^T\Omega^{-1}\Delta}\\
    &\bb{d(y}) = (\bb{y-\mu})^T\bb{\Omega}^{-1}(\bb{y-\mu})\\
    &\nu' = \nu + p\
\end{align*}

\noindent
In addition, the  $p$-dimensional $t$-PDF (i.e., $    t_{p,\nu}(\boldsymbol{y}|\boldsymbol{\mu,\Omega})$) used in Eq (3), can be calculated as follows. 

\begin{multline}
    \label{(2)}
    t_{p,\nu}(\boldsymbol{y}|\boldsymbol{\mu,\Omega}) = \frac{\Gamma(\frac{\nu + p}{2})}{\Gamma(\frac{\nu}{2})(\nu\pi)^{p/2}}|\boldsymbol{\Omega}|^{-1/2}\\ \left\{1 + \frac{(\boldsymbol{y} - \boldsymbol{\mu})^T\boldsymbol{\Omega}^{-1}(\boldsymbol{y} - \boldsymbol{\mu})}{\nu}^{-(\nu + p)/2}\right\}
\end{multline}

\noindent
In  Eq.~\eqref{(2)}, $\bb{y}$ is a random $p$-dimensional vector, $\bb{\mu}$ is the $p\times 1$ mean vector, $\bb{\Omega}$ is the $p\times p$ positive-definite symmetrical covariance matrix, $\nu$ is the S-DoF, and $\Gamma(.)$ denotes the gamma function. 

It should be highlighted that there is no analytical closed-form formulation available for $t$-CDF in Eq.~\eqref{(1)}  \cite{57}. Instead, it is estimated using direct integration of $t$-PDF or numerical approximations. In this paper, for estimating the $t$-CDF, the numerical approximation proposed by Genz et al., \cite{29} is used.  

For fitting a mixture model of multivariate skew-$t$ distributions on a dataset, an MLE method would be needed to maximize the fitness of the model (i.e., \emph{log-likelihood})  over the given dataset by iteratively optimizing the defining parameters of the kernels. For this process on $n$ observations (datapoints), the  log-likelihood  can be calculated using Eq.~\eqref{(5)}.

\begin{equation}
    \label{(5)}
    l(\boldsymbol{\Theta},\boldsymbol{y}) = \sum^{n}_{j = 1}\log\left(\sum^{g}_{i = 1}\omega_if_{p,\nu_i}(\boldsymbol{y}_j|\boldsymbol{\mu}_i,\boldsymbol{\Sigma}_i,\boldsymbol{\Delta}_i)\right)
\end{equation}

\noindent
In order to maximize this value in the MLE process, an EM algorithm is proposed in the literature \cite{16} to iteratively update the defining parameters until the convergence of the log-likelihood value. The EM algorithm contains two steps:

 \begin{enumerate}
     \item \emph{E-Step}: In this step, the expected value of the mixture model's log-likelihood will be computed given the current defining parameters of the model using Eq.~\eqref{(6)}.
     \begin{equation}
    \label{(6)}
    Q(\bb{\Theta}|\hat{\bb{\Theta}}^{(k)}) = E(l(\boldsymbol{\Theta},\boldsymbol{y})|\bb{y},\hat{\bb{\Theta}}^{(k)})
\end{equation}
 In Eq.~\eqref{(6)}, the term $\hat{\bb{\Theta}}^{(k)}$ denotes the set of mixture model's defining parameter at EM-algorithm's $k^{th}$ iteration, and $E(.)$ is the expectation operator.
     \item \emph{M-Step}: In this step, the $Q$-function calculated in the E-Step will be maximized as in Eq.~\eqref{(7)}. 
     \begin{equation}
    \bb{\Theta}^{(k+1)} = \argmax_{\bb{\Theta}} Q(\bb{\Theta}|\bb{\Theta}^{(k)})
    \label{(7)}
    \end{equation}
%     During this process, for each defining parameter, the updated value is calculated by solving the differentiation of the $Q$-function with respect to that parameter when equal to zero.
     
 \end{enumerate}

In Section \ref{erm}, we will analyze an inherent issue with the EM algorithm for training multivariate skew-$t$ mixture models.

\section{Expectation-Regularization-Maximization Algorithm} \label{erm}

% > Explain kind of similar to the intro >> What is the problem?? If there is any citation that mentions the non-optimality, non-convex cite.. if not, explain briefly what the problem is ?? SDOF diverge since there is not guarantee for convergence and also since divergence of S-DOF push the algorithm>> normal....
% To visualize the problem we run a simulation on 3 DOF (?? is it synthetic??) as can be seen eventhough one of the kernel is converging theother is diverging with increase of the .... thus the corrsponding kernel is going to be normal... 
% >> In section?? we show tht this behavior would result in low Loglikelihood fitting...

%In the literature some heuristic fixes have been implemented by not updating the sDoF when it is diverging which indeed would be counterproctive and would ... optimality... For example .... A and B

\begin{figure}[h]
    \centering
    \includegraphics[width=0.9\columnwidth]{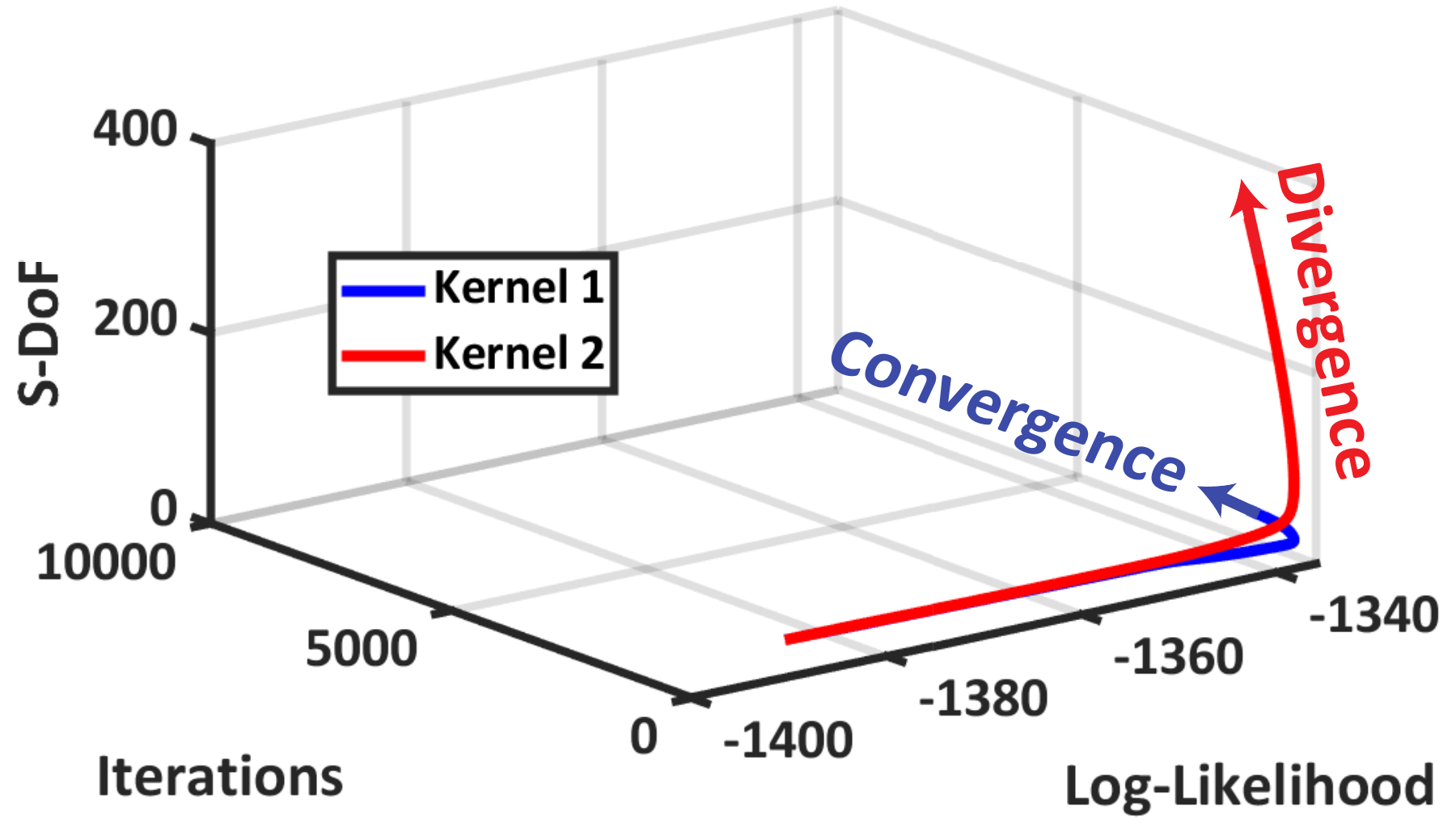}
    \caption{Example of S-DoF explosion when fitting a FM-CFUST model over AIS 2D dataset (discussed in Section \ref{datasets})}
    \label{Fig_2}
\end{figure}

 Although the EM algorithm converges for the mixture models with lower complexity, such as SNMM, we observe that for more complex models such as FM-CFUST, due to non-convex characteristics of the optimization, the convergence of the S-DoF parameter is not guaranteed. An example is given in Fig.~\ref{Fig_2} for a 2D dataset with 2 clusters. It can be observed that as the number of iterations increases, the log-likelihood will converge to a certain value, while the S-DoF of one of the FM-CFUST kernels will diverge. This phenomenon is called the S-DoF explosion in this paper. 
 Based on the mathematical representation of the multivariate skew-$t$ distribution in Section II, this monotonic increase of S-DoF will force the associated cluster to become a multivariate ``skew-normal" distribution, which is inherently less flexible than the multivariate skew-$t$ counterpart, affecting the generalizability (and robustness to outliers) by reaching a sub-optimal fit over the dataset. 
 . %This underlying behavior shows the great possibility of overfitting the mixture model using the existing EM algorithm, hence, reaching a sub-optimal fit over the dataset. 

% The divergence of the S-DoF can also result in the software’s inability to run more iterations, as the S-DoF may pass the largest finite floating-point number in IEEE double precision. 

In the existing examples for fitting Skew-$t$ mixture models, this issue is suppressed by either not updating the S-DoF value in the EM algorithm iterations if it exceeds a pre-defined heuristic value (e.g., 200) \cite{35}, or manually fixing/tuning the s-DoF \cite{25}.
 This manual manipulation will impact the iterative mathematical process by blocking parts of the designed process. Thus in this work, we propose a regularization step as part of the optimization process to theoretically stabilize the S-DoF and avoid heuristic fixes while augmenting the generalizability of the proposed framework for unseen data points.

% In this paper for the first time we propose an ERM algorithm that takes into account the divergence of the SDOF and will incorporate that as part of the optimization problem by modulating the cost function of the iterative optimization algorithm, to impose a converging behavior for the sDOF and scape the local minma which iteratively pushes the kernels to attain a normal distribution shape. For this purpose, we propose ERM algorithm as explain below ... .. 

To avoid the S-DoF explosion that can result in local minima generated by a normal or skew-normal kernel, a regularization step is implemented as an intermediate step between the E-Step and M-Step for the mixture of multivariate skew-$t$ distributions with full skewness matrices. In this step, which henceforth will be addressed as \emph{R-Step}, the expected value of the log-likelihood will be regulated using a monotonically increasing radially unbounded regularization function of S-DoF. We name the resulting iterative algorithm as \emph{Expectation-Regularization-Maximization (ERM)} algorithm, and the following equations represent the mathematical expressions of each step for the proposed ERM method:
\begin{enumerate}
    \item \emph{E-Step}: 
    \begin{equation}
    \label{(8)}
    Q(\bb{\Theta}|\hat{\bb{\Theta}}^{(k)}) = E(l(\boldsymbol{\Theta},\boldsymbol{y})|\bb{y},\hat{\bb{\Theta}}^{(k)})
    \end{equation}
    \item \emph{R-Step}: 
    \begin{equation}
    \label{(9)}
    Q'(\bb{\Theta}|\hat{\bb{\Theta}}^{(k)}) = Q(\bb{\Theta}|\hat{\bb{\Theta}}^{(k)}) - \bb{\alpha}(\bb{\nu}^{(k)})
    \end{equation}
    \item \emph{M-Step}: 
    \begin{equation}
    \label{(10)}
    \bb{\Theta}^{(k+1)} = \argmax_{\bb{\Theta}} Q'(\bb{\Theta}|\bb{\Theta}^{(k)})
    \end{equation}
\end{enumerate}

 \noindent
 In Eq.~\eqref{(9)}, $\bb{\alpha}(\bb{\nu}^{(k)})$ is the regularization function, and in this study, we define it as $\bb{\alpha}(\bb{\nu}^{(k)}) = \bb{B} \bb{\nu}^{(k)}$, in which $\bb{B}$ is a $g \times g$ diagonal matrix such that $\bb{B} = diag(\bb{\beta})$ where $\bb{\beta} = [\beta_1, \beta_2, \dots, \beta_g]$, and $\bb{\nu}^{(k)} = [\nu_1^{(k)}, \nu_2^{(k)}, \dots, \nu_g^{(k)}]^T$. We refer to the $\bb{\beta}$ parameter as the first order \emph{penalty vector}. We refer to the mixture model resulting from ERM algorithm as \emph{finite mixture of regulated skew-$t$} kernels. The full algorithmic representation of the proposed ERM method is given in Algorithm~\ref{alg_1}.

\subsection{E-Step} \label{e_step}
In order to fit an optimal multivariate skew-$t$ mixture model (based on the MLE process) we should find the solution that maximizes the expected value of the log-likelihood (i.e., the $Q$-function). To calculate this expected value, it is needed to denote the log-likelihood in terms of the latent random variables of the hierarchical representation of the multivariate skew-$t$ mixture model \cite{47} so that the likelihood of each observation is treated separately and independently of other observations in the dataset. This probabilistic representation is explained in Appendix \ref{app_a}.
In order to compute the $Q$-function given the latent variables, five conditional expected values should be computed which will be used thereafter in the M-Step. These conditional expectations are denoted as follows.

\begin{align*}
    &z_{ij}^{(k)} = E_{\bb{\Theta}^{(k)}}(\bb{Z}_{ij}|\bb{y}_j)\\
    &e_{1,ij}^{(k)} = E_{\bb{\Theta}^{(k)}}(\log(W_j)|\bb{y}_j, z_{ij} = 1)\\
    &e_{2,ij}^{(k)} = E_{\bb{\Theta}^{(k)}}(W_j|\bb{y}_j, z_{ij} = 1)\\
    &\bb{e}_{3,ij}^{(k)} = E_{\bb{\Theta}^{(k)}}(W_j\bb{U}_j|\bb{y}_j, z_{ij} = 1)\\
    &\bb{e}_{4,ij}^{(k)} = E_{\bb{\Theta}^{(k)}}(W_j\bb{U}_j\bb{U}_j^T|\bb{y}_j, z_{ij} = 1)\\
\end{align*}

\noindent
In the aforementioned probabilistic expressions, $E_{\bb{\Theta}^{(k)}}$ is the expectation operator given the estimated set of parameters $\bb{\Theta}^{(k)}$ at $k^{th}$ iteration. Although $e_{1}$, $\bb{e}_{3}$, and $\bb{e}_{4}$ are originally intractable expectations and cannot be written in closed form, by using one-step-late (OSL) approach \cite{31} for $e_1$ in Eq.~\eqref{(17)}, and expressing $\bb{e}_{3}$ and $\bb{e}_{4}$ in terms of the first and second moments of the truncated $t$-distribution (See Appendix B) respectively \cite{32,33}, the E-step can be executed in the closed form using the expressions in Eqs.~\eqref{(16)}-\eqref{(20)}.

\begin{equation}
    \label{(16)}
    z_{ij}^{(k)} = \frac{\omega_i^{(k)}f_{p,\nu_i^{(k)}}(\boldsymbol{y}_j^{(k)}|\boldsymbol{\mu}^{(k)}_i,\boldsymbol{\Sigma}^{(k)}_i,\boldsymbol{\Delta}^{(k)}_i)}{\sum^{g}_{i = 1}\omega_i^{(k)}f_{p,\nu_i^{(k)}}(\boldsymbol{y}_j^{(k)}|\boldsymbol{\mu}^{(k)}_i,\boldsymbol{\Sigma}^{(k)}_i,\boldsymbol{\Delta}^{(k)}_i)}
\end{equation}

\begin{multline}
    \label{(17)}
    e_{1,ij}^{(k)} = e_{2,ij}^{(k)} - \log\bigg(\frac{\nu_i^{(k)}+d_i^{(k)}(\bb{y}_j)}{2}\bigg)\\ - \bigg(\frac{\nu_i^{(k)}+p}{\nu_i^{(k)}+d_i^{(k)}}\bigg) -\psi(\frac{\nu_i^{(k)}+p}{2})
\end{multline}

\begin{equation}
    \label{(18)}
    e_{2,ij}^{(k)} = \bigg(\frac{\nu_i^{(k)}+p}{\nu_i^{(k)} + d_i^{(k)}(\bb{y}_j)}\bigg)\frac{T_{p,\nu_i^{(k)} + p + 2}(\bb{y}_{2,j}^{(k)}|\bb{0},\bb{\Lambda}_i^{(k)})}{T_{p,\nu_i^{(k)} + p}(\bb{y}_{1,j}^{(k)}|\bb{0},\bb{\Lambda}_i^{(k)})} 
\end{equation}

\begin{equation}
    \label{(19)}
    \bb{e}_{3,ij}^{(k)} = e_{2,ij}^{(k)}E_{\bb{\Theta}^{(k)}}(\bb{U}_{ij}|\bb{y}_j)
\end{equation}

\begin{equation}
    \label{(20)}
    \bb{e}_{4,ij}^{(k)} = e_{2,ij}^{(k)}E_{\bb{\Theta}^{(k)}}(\bb{U}_{ij}\bb{U}_{ij}^T|\bb{y}_j)
\end{equation}

\noindent
In Eq.~\eqref{(17)}, $\psi(.)$ is the Digamma function, and in Eq.~\eqref{(18)}, for $\bb{y}_{1,j}$ and $\bb{y}_{2,j}$ we have:
\begin{align*}
    &\bb{y}_{1,j} = \bb{c}_i(\bb{y}_j)\sqrt{\frac{\nu^{(k)}+p}{\nu^{(k)}+d^{(k)}(\bb{y}_j)}}\\
    &\bb{y}_{2,j} = \bb{c}_i(\bb{y}_j)\sqrt{\frac{\nu^{(k)}+p+2}{\nu^{(k)}+d^{(k)}(\bb{y}_j)}}\\
\end{align*}

\noindent
In Eq.~\eqref{(19)} and Eq.~\eqref{(20)}, the latent variable $\bb{U}_{ij}$ is distributed by truncated $t$ distribution denoted as:

\begin{align*}
    &\bb{U}_{ij}|\bb{y}_j \sim tt_{p,\nu^{(k)}+p+2}\bigg(\bb{c}_i^{(k)}(\bb{y}_j), \Big(\frac{\nu^{(k)}+d_i^{(k)}(\bb{y}_j)}{\nu^{(k)}+p+2}\Big)\bb{\Lambda}_i^{(k)}; \mathbb{R}_p^{+}\bigg)\\
\end{align*}

\noindent
where $p$ indicates the dataset dimensionality, $(\nu^{(k)}+p+2)$ is the S-DoF, $\bb{c}_i^{(k)}(\bb{y}_j)$ is the mean, $\Big(\frac{\nu^{(k)}+d_i^{(k)}(\bb{y}_j)}{\nu^{(k)}+p+2}\Big)\bb{\Lambda}_i^{(k)}$ is the covariance, and $\mathbb{R}_p^{+}$ is the $p$-dimensional truncation hyperplane (that indicates that the probability only exists in the real positive $p$-dimensional hyperspace). The details on how to compute the first and second moments of truncated $t$ distribution, i.e., $E_{\bb{\Theta}^{(k)}}(\bb{U}_{ij}|\bb{y}_j)$ and $E_{\bb{\Theta}^{(k)}}(\bb{U}_{ij}\bb{U}_{ij}^T|\bb{y}_j)$, are discussed in Appendix \ref{app_b}.

\subsection{R-Step} \label{r_step}
To avoid S-DoF explosion, we regulate the $Q$-function in the R-Step using the monotonically increasing radially unbounded regularization function of S-DoF (i.e., $\bb{B} \bb{\nu}^{(k)}$), as seen in Eq.~\eqref{regulation}. This regularization will act as a penalty term when combined with the expected value of the log-likelihood and thus penalizes the sub-optimal excessive increase in the S-DoF values throughout the course of the ERM optimization, keeping the model as a multivariate skew-$t$ mixture. %Moreover, by increasing the elements of $\Beta$, the penalty effect would be more impactful in regularizing the S-DoF values.

\begin{equation}
    \label{regulation}
    Q'(\bb{\Theta}|\hat{\bb{\Theta}}^{(k)}) = Q(\bb{\Theta}|\hat{\bb{\Theta}}^{(k)}) - \bb{B} \bb{\nu}^{(k)}
\end{equation}

The resulting regulated $Q'$-function will be used in the M-Step of the proposed ERM algorithm for multivariate skew-$t$ based mixture models. This will inject the regulation needed for the ERM algorithm to modulate the updating dynamics of S-DoF and to avoid the S-DOF Explosion which would result in suboptimal normal or skew-normal clusters in the final mixture model. The effect of the penalty vector on the S-DoF convergence during the ERM algorithm iterations is discussed in Section \ref{initial}, and the results of our study showed the superior performance of the proposed method in comparison to the existing counterparts.

\subsection{M-Step} \label{m_step}
For the M-step, using $\bb{\Theta}^{(k)}$ and the calculations in the E-step, the parameters for the mixture model are updated for the $(k+1)^{th}$ iteration to ensure global maximization of the $Q'$-function, as seen in Eq.~\eqref{(10)}. This expression yields the following steps in Eqs.~\eqref{(21)}-\eqref{(24)} for updating the mixing proportion, mean vector, covariance matrix, and skewness matrix for each cluster. 

\begin{equation}
    \label{(21)}
    \omega_i^{(k+1)} = \frac{1}{n}\sum^n_{j = 1}z_{ij}^{(k)}
\end{equation}

\begin{equation}
    \label{(22)}
    \bb{\mu}_i^{(k+1)} = \frac{\sum^n_{j=1}z_{ij}^{(k)}e_{2,ij}^{(k)}\bb{y}_j - \bb{\Delta}_i^{(k)}\sum^n_{j=1}z_{ij}^{(k)}\bb{e}_{3,ij}^{(k)}}{\sum^n_{j=1}z_{ij}^{(k)}e_{2,ij}^{(k)}}
\end{equation}

\begin{equation}
    \label{(23)}
    \bb{\Delta}_i^{(k+1)} = \bigg(\sum_{j=1}^n z_{ij}^{(k)}(\bb{y}_j - \bb{\mu}_i^{(k+1)})\bb{e}_{3,ij}^{(k)}\bigg)\bigg(\sum^n_{j=1}z_{ij}^{(k)}\bb{e}_{4,ij}^{(k)}\bigg)^{-1}
\end{equation}

\begin{multline}
    \label{(24)}
    \bb{\Sigma}_i^{(k+1)} = \bigg(\sum_{j=1}^n z_{ij}^{(k)}\Big[e_{ij}^{(k)}(\bb{y}_j - \bb{\mu}_i^{(k+1)})(\bb{y}_j - \bb{\mu}_i^{(k+1)})^T   \\
    -\bb{\Delta}_i^{(k+1)}\bb{e}_{3,ij}^{(k)}(\bb{y}_j - \bb{\mu}_i^{(k+1)})^T\\
    -(\bb{y}_j - \bb{\mu}_i^{(k+1)})\bb{e}_{3,ij}^{(k)}\bb{\Delta}_i^{(k+1)}\\
    + \bb{\Delta}_i^{(k+1)}\bb{e}_{4,ij}^{(k)}\bb{\Delta}_i^{(k+1)T}\Big]\bigg)
    \bigg(\sum_{j=1}^n z_{ij}^{(k)}\bigg)^{-1}
\end{multline}

\noindent
Ultimately, for updating the S-DoF for each cluster, the maximization process is done by solving Eq.~\eqref{(25)} for $\nu_i^{(k+1)}$, which is the derivative of the $Q'$-function with respect to the S-DoF when equated to zero.

\begin{multline}
     \log(\frac{\nu_i^{(k+1)}}{2})-\psi(\frac{\nu_i^{(k+1)}}{2}) + 1 \\ - \frac{\sum_{j=1}^n z_{ij}^{(k)}(e_{2,ij}^{(k)} - e_{1,ij}^{(k)})}{\sum_{j=1}^n z_{ij}^{(k)}} - \frac{\partial \alpha(\nu_i^{(k+1)})}{\partial \nu_i^{(k+1)}} = 0
     \label{(25)}
\end{multline}

Algorithm~\ref{alg_1} shows how this approach can be implemented to train this distribution based on a given dataset $\bb{y}$.

\begin{algorithm}[h]
    \caption{FiMReS$t$ Distribution}
     \begin{algorithmic}[1]
     \renewcommand{\algorithmicrequire}{\textbf{Input:}}
     \renewcommand{\algorithmicensure}{\textbf{Output:}}
     \REQUIRE $\bb{y},g$
     \ENSURE  $\bb{\mu},\bb{\Delta},\bb{\Sigma},\bb{\nu},\bb{\omega},l$
     \\ \textit{Initialisation} :  $\bb{\mu}^{(0)},\bb{\Delta}^{(0)},\bb{\Sigma}^{(0)},\bb{\nu}^{(0)},\bb{\omega}^{(0)}$ \cite{16}
      \STATE Set $\delta\bb{\nu}$, $\delta l$, $\bb{\alpha(\nu)}$
      \STATE $k \leftarrow 1$, $\bb{t}_1 \leftarrow \delta\bb{\nu}+1$, $\bb{t}_2 \leftarrow \delta l+1$
      \WHILE{($|\bb{t}_1| > \delta\bb{\nu}$) and ($|\bb{t}_2| > \delta l$)}
      \STATE Compute: $Z$, $e_1^{(k)}$, $e_2^{(k)}$, $\bb{e}_3^{(k)}$, $\bb{e}_4^{(k)}$ using Eqs.~\eqref{(16)}-\eqref{(20)}
      \STATE Compute: $\bb{\omega}^{(k)},\bb{\mu}^{(k)},\bb{\Delta}^{(k)},\bb{\Sigma}^{(
      k)}$ using Eqs.~\eqref{(21)}-\eqref{(24)}
      \STATE Solve Eq.~\eqref{(25)} to update $\bb{\nu}^{(k)}$
      \STATE Compute: $l^{(k)} = \sum^{n}_{j = 1}\log\left(\sum^{g}_{i = 1}\omega_if_{p,\nu_i}(\bb{y}|\bb{\mu}_i,\bb{\Sigma}_i,\bb{\Delta}_i)\right)$
      \IF {($k > 2$)}
      \STATE $\bb{t}_1 \leftarrow (\bb{\nu}^{(k)} - \bb{\nu}^{(k-1)})$
      \STATE $\bb{t}_2 \leftarrow (l^{(k)} - l^{(k-1)})$
      \ENDIF
      \STATE $k \leftarrow k + 1$
      \ENDWHILE
     \RETURN $\bb{\mu},\bb{\Delta},\bb{\Sigma},\bb{\nu},\bb{\omega},l$
     \end{algorithmic}
     \label{alg_1}
\end{algorithm}

\section{Experimental Results} \label{results}

In this section, we present the results of the FiMReS$t$ implementation and the comparison of this new form of mixture modeling with previously proposed techniques (namely GMM, SNMM, STMM, and FM-CFUST) on several datasets (including experimental and synthetic). Also, we evaluate the dependency of the proposed model on the initial values of S-DoF in addition to the effect of the penalty vector on the S-DoF convergence. 

\subsection{Datasets} \label{datasets}

\subsubsection{Australian Institute of Sport (AIS) Dataset}
In this work, we utilize the AIS dataset, which is used frequently in the literature (e.g., \cite{16,22}) due to the size and inherent asymmetry that would provide a valuable framework for evaluating the performance of various probabilistic mixture models.   This dataset contains 13 different measurements, such as body fat and body mass index, from 202 male and female athletes \cite{36}. In this study, we extract two sub-datasets.%, which are the same datasets that have been used for performance evaluation in \cite{16} and \cite{22}.
The first (that will be referred to as the \textit{AIS 2D dataset} in the rest of this paper) is a bivariate set containing height in centimeters and percentage of body fat. The second (that will be referred to as the \textit{AIS 3D dataset} in the rest of this paper) is a trivariate (3D) set that contains body mass index, lean body mass, and body fat.

\subsubsection{Surface Electromyography (sEMG) Neck Muscle Coherence Dataset}

We test the performance of the FiMReS$t$ model on three different datasets extracted from the muscle activity frequency coherence database acquired by our team at NYU using sEMG sensors on perilaryngeal and cranial muscles \cite{56,60}. This database contains 48 maximum (over frequency) coherence values (ranging from $0$ to $1$) of sEMG muscle activation signals of 14 muscles during various vocal tasks, highlighting co-modulated muscle activations. We extract three different 2D datasets (named here as \textit{Part 1}, \textit{Part 2}, and \textit{Part 3}), each composed of 2 dimensions representing two pairs of muscles as explained in the following:

\begin{enumerate}
    \item \textit{Part 1}: The maximum coherence between the left and right Sternohyoid muscle; and The Maximum coherence between  left-lower and left-upper Sternocleidomastoid muscles. 
    \item \textit{Part 2}: The maximum coherence between the left front and back Trapezius; and The maximum coherence between the right front and back Trapezius.
    \item \textit{Part 2}: The maximum coherence between the left and right Masseter; and The maximum coherence between the left and right Omohyoid.
\end{enumerate}

\subsubsection{Synthetic Dataset}

In addition to the real-world experimental datasets, mentioned above, in this paper, we also devised a unitless synthetic dataset with embedded outliers by scattering two random asymmetrically distributed points in a $100\times100$ square, with a diagonal separation and outliers at the ends of the separating lines (seen in Fig.~\ref{Fig_7}(a)). We refer to this dataset as \textit{Synthetic Dataset} henceforth.

\subsection{Comparative Study} \label{implementation}

\begin{figure}
    \centering
    \includegraphics[width=0.9\columnwidth]{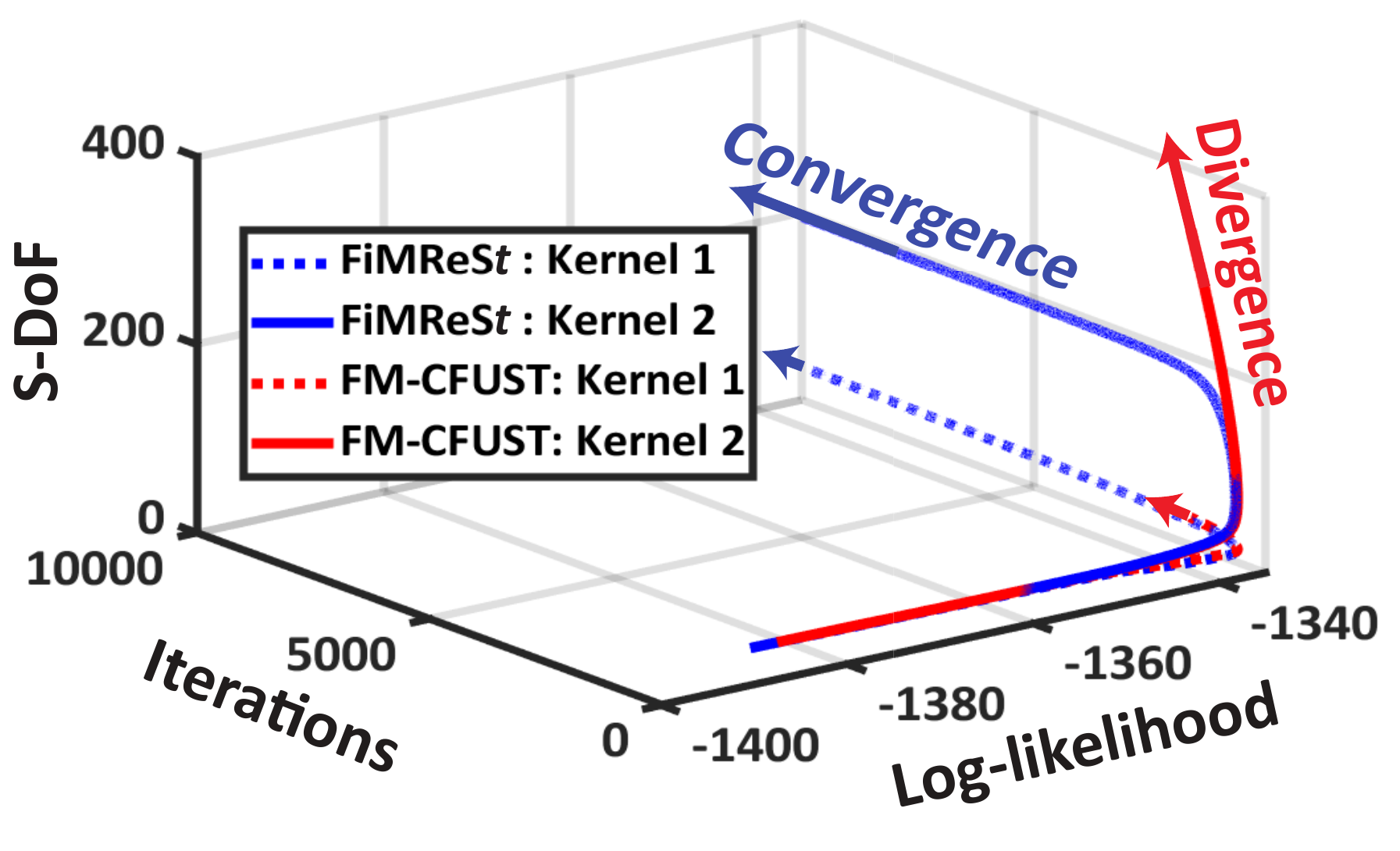}
    \caption{3D plot of S-DoF and Log-likelihood vs. iteration for fitting the FiMReS$t$ and FM-CFUST over the AIS 2D dataset. It can be seen that, unlike FM-CFUST, FiMReS$t$ will preserve the converging behavior of the S-DoFs.}
    \label{Fig_3}
\end{figure}

Fig~\ref{Fig_3} shows the stabilized behavior of the S-DoF for the proposed FiMReS$t$ over the course of ERM iterations, compared to the S-DoF explosion in fitting FM-CFUST (from the literature) using the EM algorithm. In this figure, the blue lines show the S-DoF of the kernels in the proposed FiMReS$t$ model, and the red lines show the S-DoF of the kernels in the  FM-CFUST model.  For the FiMReS$t$ model in this case, the penalty vector is chosen to be $\bb{\beta} = \begin{bmatrix} 0.000005 \ \ 0.000005  \end{bmatrix}^\intercal$. As expected, we can see that by using the proposed Algorithm~\ref{alg_1}, the S-DoF explosion will be avoided, and the kernels will converge as the number of iterations increases.

\begin{table} 
\renewcommand{\arraystretch}{1.5}
    \centering
    \caption{Log-Likelihood of various mixture models over the AIS 2D dataset}\vspace{-0.2cm}
    \begin{tabular}{|c|c|}
     \hline
         Model &  Log-Likelihood\\  \hline \hline
         GMM & -1351.67\\  \hline
         SNMM &	-1341.12\\  \hline
         STMM &	-1340.95 \cite{16}\\  \hline
         FM-CFUST & -1335.60\\  \hline
         \textbf{FiMReS$t$}	& \textbf{-1335.20}  \\ \hline
    \end{tabular}
    \label{tab_1}
\end{table}

\begin{figure}
    \centering
    \includegraphics[width=0.75\columnwidth]{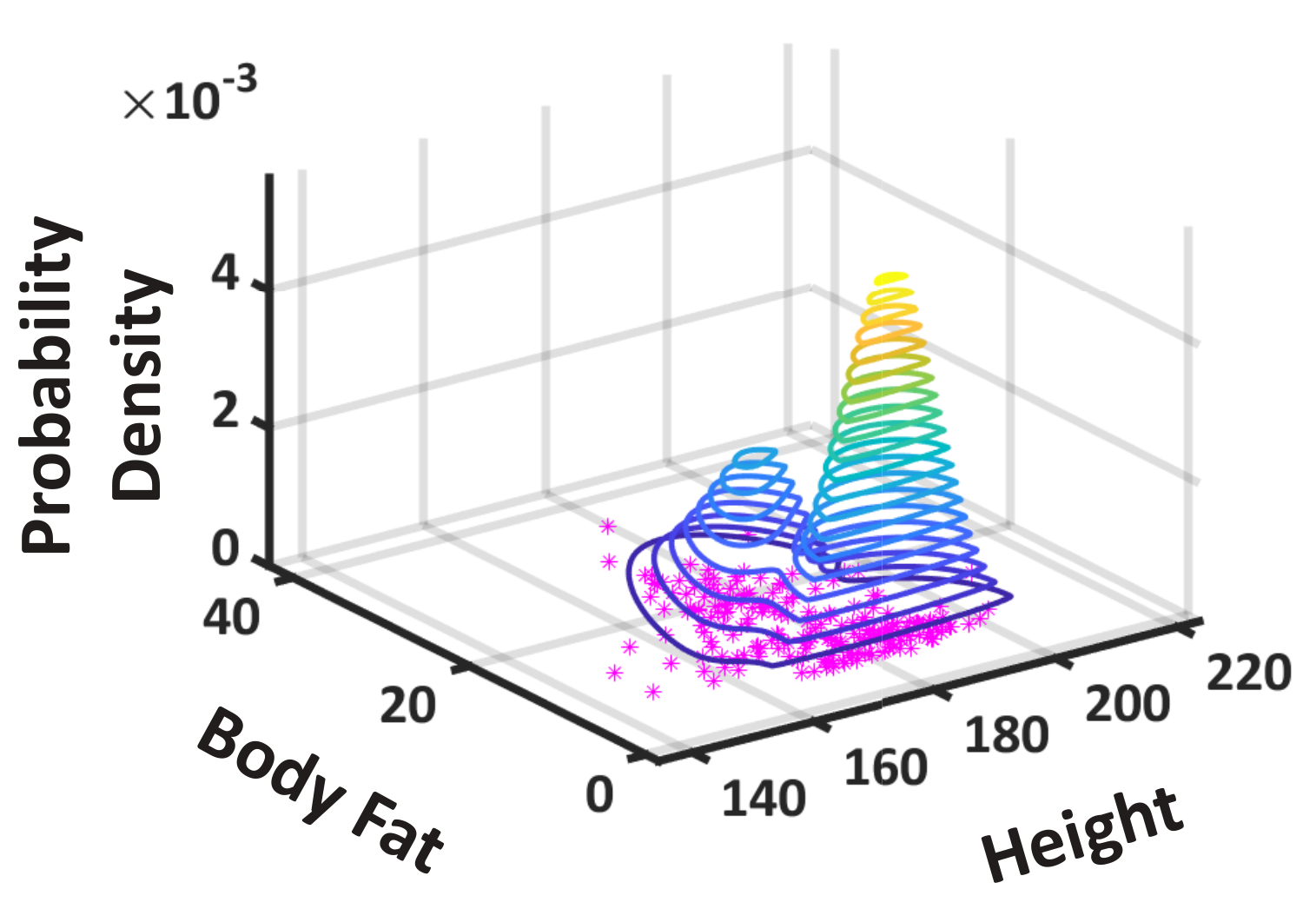}
    \caption{Probability density function visualization of the FiMReS$t$ model over AIS 2D dataset.}
    \label{Fig_4}
\end{figure}
The resulting convergence, seen in Fig.~\ref{Fig_3}, will allow the training to continue while preserving the flexibility of the multivariate skew-$t$ distribution and preventing the kernels from converging to a normal mixture distribution and preventing internal numerical instability (which would halt the training). This behavior will allow the mixture model to continue the exploration and learning of the underlying physics of the dataset space boosting optimality and generalizability. The log-likelihood comparison from fitting various mixture models over AIS 2D dataset is given in Table~\ref{tab_1}. As can be seen, FiMReS$t$ secures the highest log-likelihood when compared with all other mixture models which again supports the performance of the proposed algorithm. Fig~\ref{Fig_4} shows how the kernels of the FiMReS$t$ model fit over AIS 2D.

\begin{figure}[h]
    \centering
    \includegraphics[width=0.95\columnwidth]{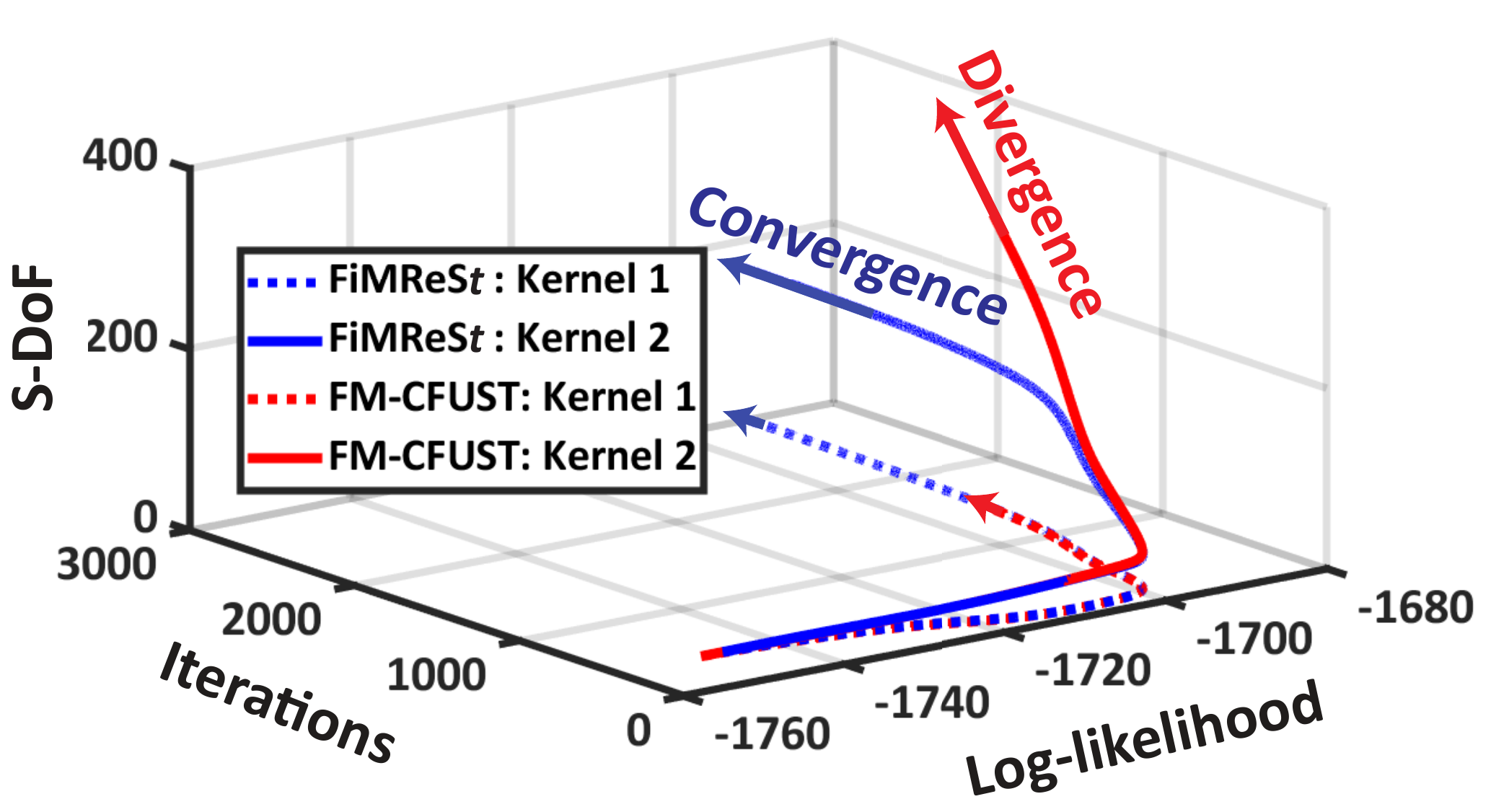}
    \caption{Comparison of S-DoF behavior during fitting iterations of FiMReS$t$ and FM-CFUST over AIS 3D dataset}
    \label{Fig_5}
\end{figure}

\begin{figure}[h]
    \centering
    \includegraphics[width=0.9\columnwidth]{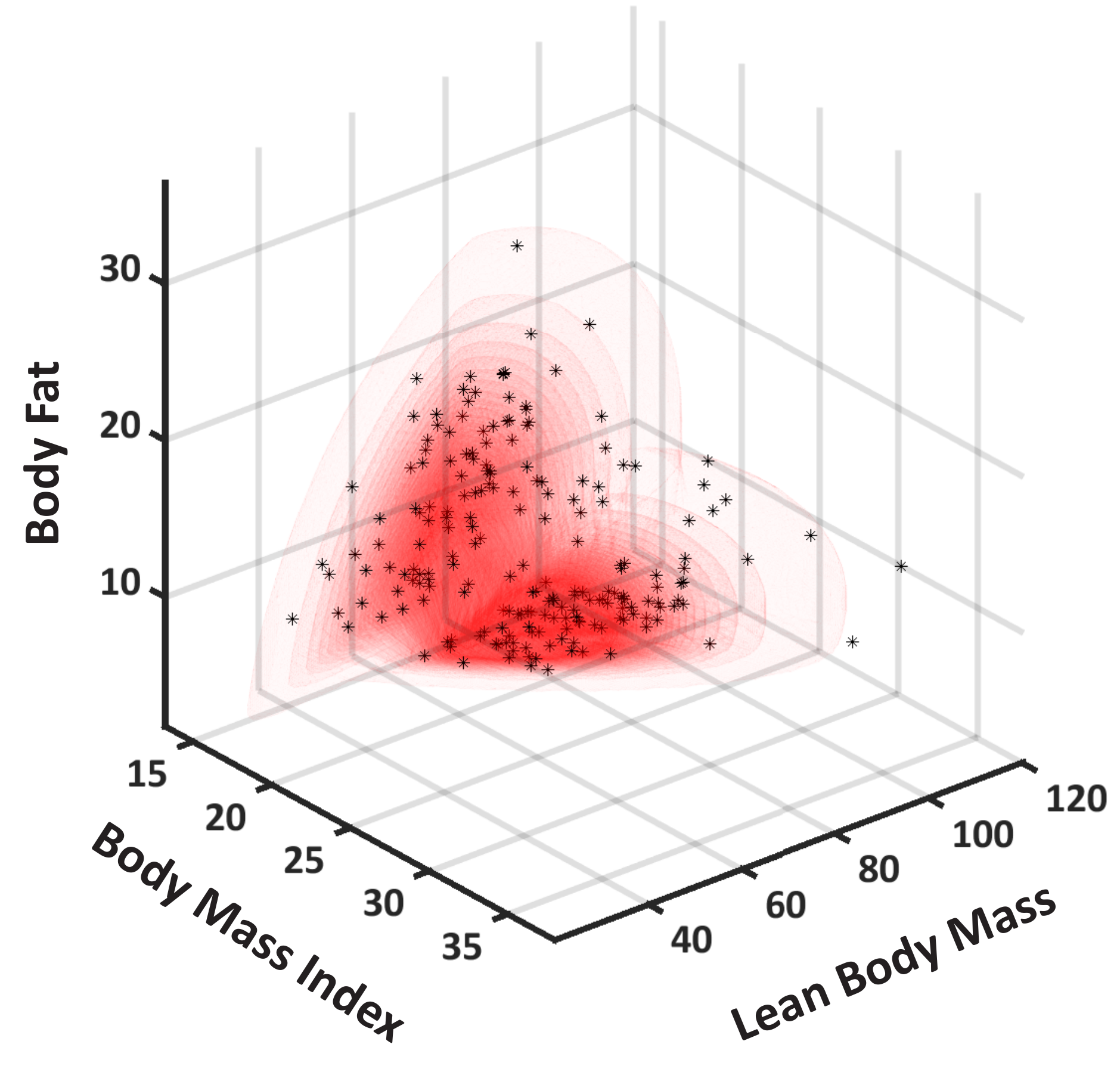}
    \caption{Probability density function visualization of the FiMReS$t$ kernels over AIS 3D dataset. Higher probabilities are shown by darker red color in the visualization. The outermost 3D layer indicates 1\% of the maximum probability produced by the FiMReS$t$ model in the 3D space.}
    \label{Fig_6}
\end{figure}

Fig.~\ref{Fig_5} is about the  AIS 3D dataset to show the performance of the model on a larger dimension. The figure shows the S-DoF behavior comparison during the proposed FiMReS$t$ and FM-CFUST modeling. Similar to Fig.~\ref{Fig_3}, the lines in Fig.~\ref{Fig_5} show the S-DoF values of the FiMReS$t$ and FM-CFUST throughout the iterative training process. The same observation can be made in Fig.~\ref{Fig_5} regarding the AIS 3D dataset case that by avoiding the divergence of the S-DoF values, the FiMReS$t$ model will continue the fitting process and achieve higher log-likelihood in comparison. (refer to Table~\ref{tab_2}). Fig.~\ref{Fig_6} depicts the visualization of the FiMReS$t$'s kernels fitted on the AIS 3D dataset. The trained parameters of the FiMReS$t$ model over the AIS 3D dataset are given in Table~\ref{tab_3}.

\begin{table}[h]
\renewcommand{\arraystretch}{1.5}
    \centering
        \caption{Log-Likelihood of the various mixture modeling techniques over AIS 3D dataset}\vspace{-0.2cm}
    \begin{tabular}{|c|c|}
     \hline
         Model &  Log-Likelihood\\  \hline \hline
         GMM & -1747.20\\  \hline
         SNMM &	-1726.17\\  \hline
         STMM &	-1725.01\\  \hline
         FM-CFUST & -1700.17 \cite{22}\\  \hline
         \textbf{FiMReS$t$}	& \textbf{-1692.08}  \\ \hline
    \end{tabular}
    \label{tab_2}
\end{table}

\begin{table}[!t]
\renewcommand{\arraystretch}{1.5}
    \centering
        \caption{Defining parameters of the FiMReS$t$ model given the AIS 3D dataset} \vspace{-0.2cm}
    \begin{tabular}{|c|c|c|}
     \hline
         $i$ & 1 & 2\\  \hline
         $\bb{\mu}_i$ & $\begin{bmatrix} 19.745 \\ 57.698 \\ 11.737  \end{bmatrix}$ & $\begin{bmatrix} 20.565 \\ 63.694 \\ 5.773  \end{bmatrix}$ \\  \hline
         $\bb{\Delta}_i$& 
         $\begin{bmatrix} 3.616 & -1.979 & 1.090 \\ 5.652 & -8.113 & -1.407 \\ 3.582 & -3.489 & 7.577 \end{bmatrix}$ & 
         $\begin{bmatrix} 2.792 & -0.294 & 1.170 \\ 6.544 & 4.343 & 1.021 \\ 0.198 & -0.053 & 3.779 \end{bmatrix}$ \\  \hline
         $\bb{\Sigma}_i$&
         $\begin{bmatrix} 0.046 & -0.377 & -0.116 \\ -0.377 & 7.170 & 2.411 \\ -0.116 & 2.411 & 0.816 \end{bmatrix}$ &
         $\begin{bmatrix} 0.678 & 5.733 & 0.243 \\ 5.733 & 48.535 & 2.055 \\ 0.243 & 2.055 & 0.088 \end{bmatrix}$ \\  \hline
         $\nu_i$ & 173.856 & 7.443\\  \hline
         $\omega_i$ & 0.4806 & 0.5194\\  \hline
         $\beta_i$ & 0.0001 & 0.0001\\ \hline
    \end{tabular}
    \label{tab_3}
\end{table}

\begin{figure}[h]
    \centering
    \includegraphics[width=\columnwidth]{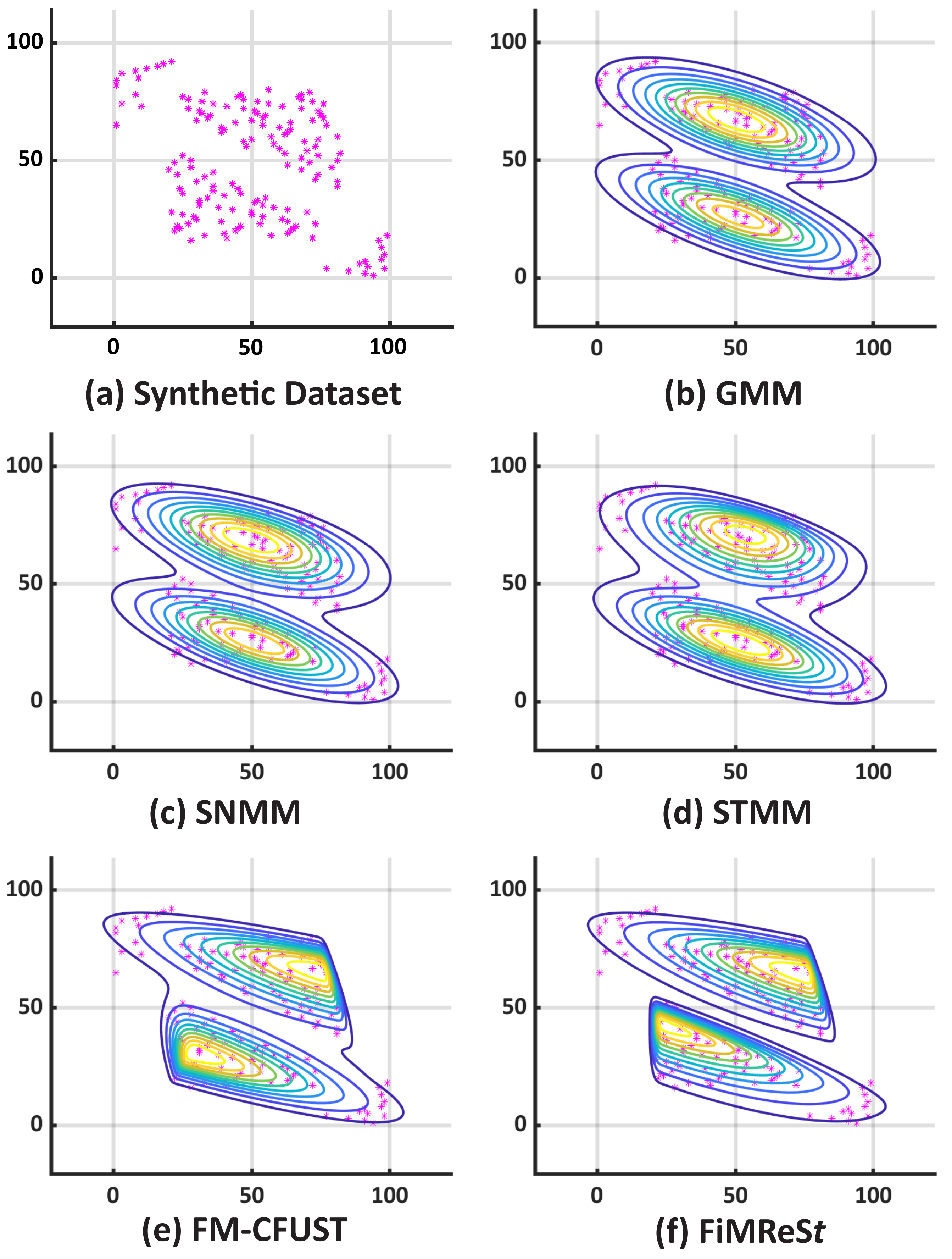}
    \caption{ Comparison of GMM, SNMM, STMM, FM-CFUST and FiMReS$t$ on the synthetic dataset}
    \label{Fig_7}
\end{figure}

Fig.~\ref{Fig_7} shows the comparison between GMM, SNMM, STMM, FM-CFUST, and the proposed FiMReS$t$ over the synthetic dataset, and Table~\ref{tab_4} enumerates the log-likelihood values of the corresponding models. In Fig.~\ref{Fig_7}(a), the synthetic dataset is visualized. In Fig.~\ref{Fig_7}(b), a GMM with two Gaussian kernels is trained on the synthetic dataset. In Fig.~\ref{Fig_7}(c), we can see how the skewness in SNMM will cause the kernels to adapt better to the formation of the data sparsity. In Fig.~\ref{Fig_7}(d), the STMM model is shown, and the difference between STMM and SNMM that is caused by changing the skew-normal clusters to skew-$t$ clusters can be seen. %It should be noted that even though the STMM can theoretically approach higher performance compared with the SNMM model, due to the S-DoF explosion, the fitting process will seize to improve the model further. As a result, the log-likelihood of the SNMM model is slightly better than the STMM model.
Fig.~\ref{Fig_7}(e) shows the FM-CFUST model trained over the synthetic dataset benefiting from the full skewness matrix (rather than diagonal) resulting in a better fit when compared with STMM.  Fig.~\ref{Fig_7}(f) depicts the behavior of the proposed FiMReS$t$ model trained on the synthetic dataset. It can be seen that by stabilizing the S-DoF parameters of the kernels in the proposed ERM algorithm, avoiding S-DoF explosion, while exploiting the full potential of the skewness matrix using the math of Skew-$t$ distribution, the FiMReS$t$ model will converge to a higher log-likelihood and secure the best fit, achieving a more conforming shape to the dataset. The kernels in the FiMReS$t$ model are showing more reach towards the outliers of the synthetic dataset, causing the clustering to be more divided in comparison to FM-CFUST and other models. 

\begin{table}[!h]
\renewcommand{\arraystretch}{1.5}
    \centering
        \caption{Log-Likelihood of the synthetic dataset given different models}\vspace{-0.2cm}
    \begin{tabular}{|c|c|}
    
     \hline
         Model &  Log-Likelihood\\  \hline \hline
         GMM & -1390.541\\  \hline
         SNMM &	-1389.482\\  \hline
         STMM &	-1389.924\\  \hline
         FM-CFUST & -1364.918\\  \hline
         \textbf{FiMReS$t$}	& \textbf{-1362.914}  \\ \hline
         
    \end{tabular}
    \label{tab_4}
\end{table}

\begin{figure*}[t!]
    \centering
    \includegraphics[width=2\columnwidth]{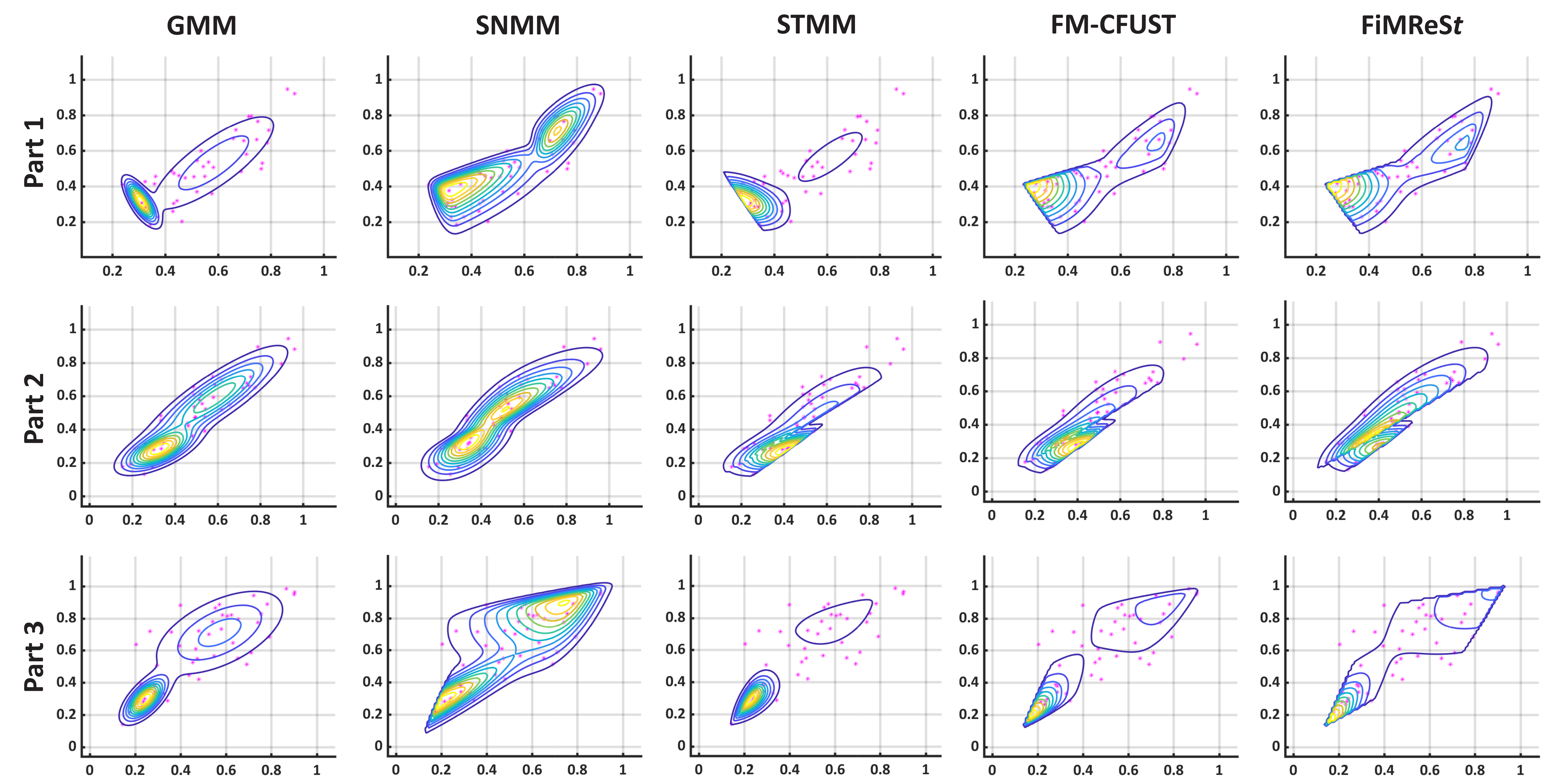}
    \caption{The comparison between GMM, SNMM, STMM, FM-CFUST, and FiMReS$t$ over coherence datasets}
    \label{Fig_8}
\end{figure*}

Fig.~\ref{Fig_8} shows the comparison of the FiMReS$t$ model with GMM, SNMM, STMM, and FM-CFUST trained on the experimental sEMG coherence dataset for all data parts 1, 2 and 3. In Fig.~\ref{Fig_8}, the rows indicate which data parts used, and columns show the visualization of the resulting clustering methods, i.e., GMM, SNMM, STMM, FM-CFUST, and FiMReS$t$, from left to right, respectively. The penalty vector for FiMReS$t$ model for all three parts of the dataset is $\bb{\beta} = \begin{bmatrix} 0.00001 \ \ \ 0.00001  \end{bmatrix}^\intercal$. In Fig.~\ref{Fig_8}, it can be observed that the FiMReS$t$ distribution will adapt more efficiently to the datasets as the generalizability of the multivariate skew-$t$ distribution is preserved during the ERM algorithm iterations and this is supported by the numerical results given in Table  Table~\ref{tab_5}. In addition to this main observation, there are other points to be considered, as explained in the following. In Fig.~\ref{Fig_8}, the difference between GMM and SNMM can be seen due to the added skewness. The SNMM model will conform to the dataset more considerably by relaxing the symmetricity assumption of the Gaussian kernels. Thereafter, the replacement of MSN with MST kernels results in a noticeable change in STMM compared to SNMM. Going one step further from STMM, by taking advantage of the full skewness matrix, the FM-CFUST model will show more  conformity to the given datasets. Finally, by implementing the proposed ERM method and avoiding the S-DoF explosion while taking into account the full skewness, the FiMReS$t$ model  further expands towards modeling the outliers in the  scattered segments of the data, resulting in a more optimal and generalizable clustering behavior and a higher log-likelihood, given in Table~\ref{tab_5}.

\begin{table}[h]
\renewcommand{\arraystretch}{1.5}
    \centering
        \caption{Log-likelihood of GMM, SNMM, STMM, FM-CFUST, and FiMReS$t$ over coherence parts 1, 2, and 3} \vspace{-0.2cm}
    \begin{tabular}{|c|c|c|c|}
    \hline
        Model & Part 1 & Part 2 & Part 3\\ \hline \hline
        GMM & 67.0987 & 64.3616 & 45.3951\\ \hline
        SNMM & 69.7102 & 64.5459 & 51.0562\\ \hline
        STMM & 70.4885 & 71.3382 & 49.5397\\ \hline
        FM-CFUST & 75.2098 & 72.4159 & 52.3720 \\ \hline
        \textbf{FiMReS$t$} & \textbf{76.5683} & \textbf{72.9244} & \textbf{58.1812} \\
        \hline
    \end{tabular}

    \label{tab_5}
\end{table}

\subsection{Initial Conditions} \label{initial}

\begin{figure}[h]
    \centering
    \includegraphics[width=0.7\columnwidth]{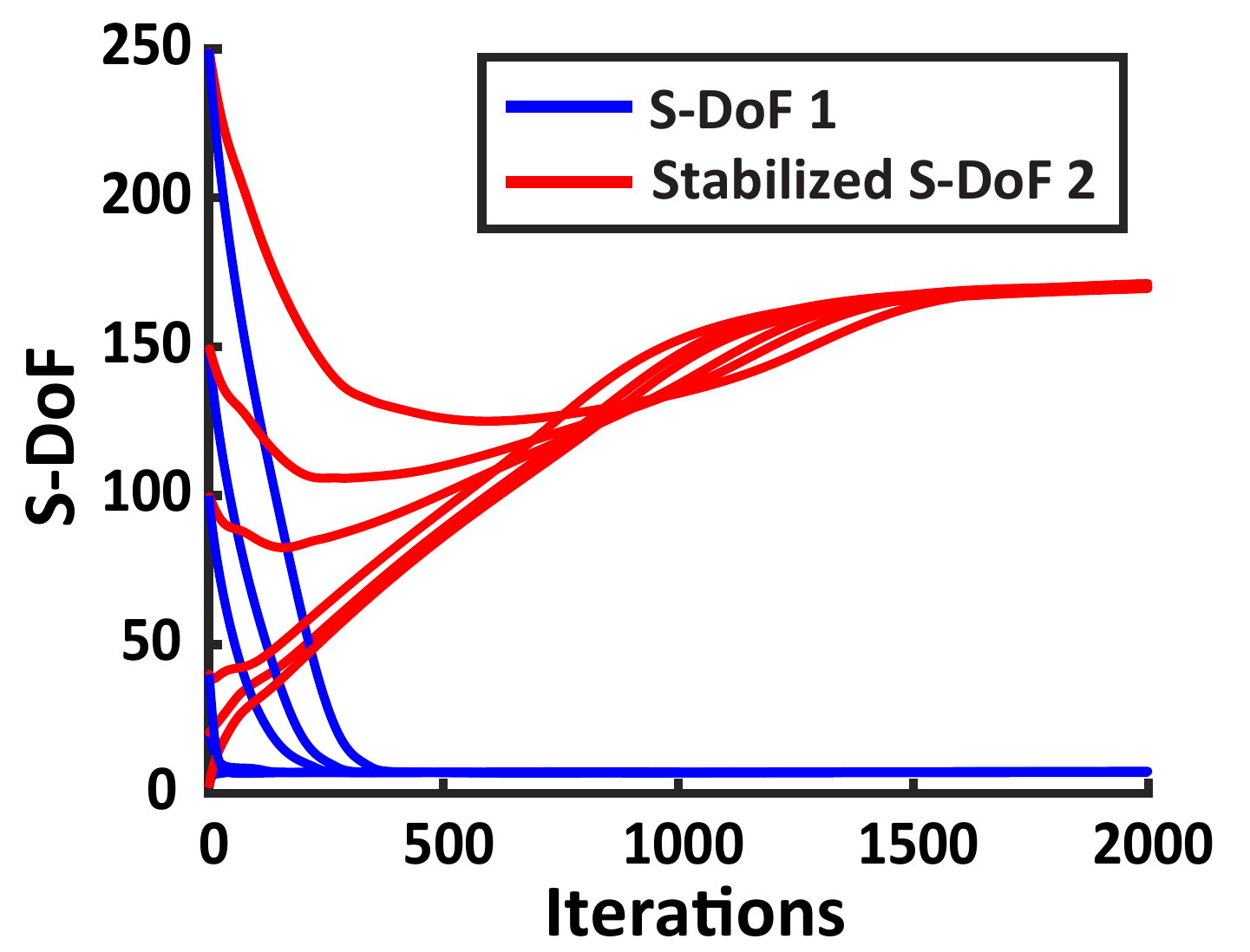}
    \caption{S-DoF convergence of the FiMReS$t$ over the AIS 3D model given different initial values. The red lines show the stabilized S-DoF.}
    \label{Fig_10}
\end{figure}

In this sub-section, the sensitivity of the proposed model, in terms of the convergence of the S-DoF, to the initial values of the S-DoF is investigated. High sensitivity would reduce the practicality of the method.  Fig.~\ref{Fig_10} shows the convergence of the S-DoF values of FiMReS$t$ model for the two clusters of the AIS 3D dataset, considering a large range of initial values for S-DoF, i.e.,  $\nu_i \in \{2,20,100,150,250\}$. We can see that the resulting FiMReS$t$ model has minimal to no dependency on the initial values of S-DoF, which corroborates the robustness of this mixture model to the S-DoF initialization. This further supports the practicality of the proposed model.

\section{Conclusion} \label{con}

In this work, we proposed an expectation-regularization-maximization algorithm to train a finite mixture of multivariate skew-$t$ distributions. The resulting mixture model is named Finite Mixture of Multivariate Regulated Skew-$t$ (FiMReS$t$) Kernels, which, unlike its counterparts in the literature, is able to stabilize the S-DoF profile during the optimization process of learning, boosting generalizability and robustness to outliers. The proposed method is evaluated through a comprehensive comparative study based on real and synthetic data. The results support the higher performance of the proposed  FiMReS$t$ model when compared with existing methods (i.e., GMM, SNMM, STMM, FM-CFUST) in terms of the log-likelihood of fit and the numerical stability of S-DoF. Additional analysis highlighting the low sensitivity of the proposed method to the initialization of S-DoF further supports the practicality of the approach.

\appendices
\section{Hierarchical Representation of the Multivariate Skew-$t$ Distribution} \label{app_a}

In the MLE process, the key assumption is that the likelihood of each observation is independent of the likelihood of other observations. Therefore, to satisfy this assumption, the multivariate skew-$t$ mixture model PDF in Eq.~\eqref{CFUST} is represented in the hierarchical format \cite{26, 47} for the ERM algorithm, as seen in Eq.~\eqref{(11)}.

\begin{equation}
    \begin{split}
        &\boldsymbol{Y}_j|\boldsymbol{u}_j,w_j,z_{ij}=1 \sim N_p(\boldsymbol{\mu}_i + \boldsymbol{\Delta}_i \boldsymbol{u}_j,\frac{1}{w_j}\boldsymbol{\Sigma}_i)\\
        &\boldsymbol{U}_j|w_j,z_{ij}=1 \sim HN_p(\boldsymbol{0},\frac{1}{w_j}\boldsymbol{I}_p)\\
        &W_j|z_{ij}=1\sim \Gamma(\frac{\nu_i}{2},\frac{\nu_i}{2})\\
        &\boldsymbol{Z}_j \sim \mathcal{M}_g(1;\omega_1,\omega_2,\dots,\omega_g)
    \end{split}
    \label{(11)}
\end{equation}

\noindent
In Eq.~\eqref{(11)}, $g$ is the number of multivariate skew-$t$ kernels in the mixture model, $j$ subscript addresses the corresponding parameter to the $j^{th}$ observed point in the dataset, such that $j = 1,\dots,n$, and $i$ subscript addresses the parameter related to the $i^{th}$ multivariate skew-$t$ kernel, where $i = 1,\dots,g$. In addition, $HN_p(.,.)$ and $\Gamma(.,.)$ denote the $p$-dimensional half-normal, and Gamma distributions, respectively. For a multivariate skew-$t$ mixture model with $g$ clusters, a $g\times n$ matrix $\bb{Z}$ is defined to be associated with how much each cluster is responsible for each data point in the set, hence $\sum^g_{i = 1} z_{ij} = 1$. This means that if $z_{ij} = 1$, $\boldsymbol{y}_j$ belongs solely to the cluster $i$. Utilizing the aforementioned definition, we can see that each random vector $\boldsymbol{Z}_j = [z_{1j}, z_{2j}, \dots, z_{gj}]^T$ is distributed by a multinomial distribution with one trial, i.e., $\boldsymbol{Z}_j \sim \mathcal{M}_g(1;\omega_1,\omega_2,\dots,\omega_g)$, and the cell probabilities $\omega_i$ are the mixing proportions of the mixture model. 

The hierarchical representation of the mixture model, utilizes random values that are unseen in the mathematical representation of the mixture model distribution in Eq.~\eqref{CFUST}, namely $\bb{Z}$, $\bb{W}$, and $\bb{U}$. These parameters are the so-called latent variables of the mixture model, and the ERM algorithm relies on the computation of these latent variables.

By expanding Eq.~\eqref{(5)}, we can decompose the log-likelihood as seen in Eq.~\eqref{(12)}.

\begin{equation}
    l(\boldsymbol{\Theta}|\boldsymbol{y,u},w,\boldsymbol{z}) = L_1(\boldsymbol{\Theta}|\boldsymbol{z}) + L_2(\boldsymbol{\Theta}|w,\boldsymbol{z}) + L_3(\boldsymbol{\Theta}|\boldsymbol{y,u},w,\boldsymbol{z})
    \label{(12)}
\end{equation}

\noindent
where each term can be re-written as a function of the defining parameters of the multivariate skew-$t$ clusters in the mixture model and the latent variables of the hierarchical distribution function of the mixture model following Eqs.~\eqref{(13)}-\eqref{(15)}.
\begin{equation}
    L_1(\boldsymbol{\Theta}|\boldsymbol{z})=\sum^g_{i=1}\sum^n_{j = 1}z_{ij}\log(\omega_i)
    \label{(13)}
\end{equation}

\begin{multline}
     L_2(\boldsymbol{\Theta}|w,\boldsymbol{z})=\sum^g_{i=1}\sum^n_{j = 1}z_{ij}\Big\{
     (\frac{\nu_i}{2})\log(\frac{\nu_i}{2}) + (\frac{\nu_i}{2} + p - 1)\log(w_j)\\ - \log\Gamma(\frac{\nu_i}{2}) - \frac{w_j}{2}\nu_i
     \Big\}    
     \label{(14)}
\end{multline}

\begin{multline}
    L_3(\boldsymbol{\Theta}|\boldsymbol{y,u},w,\boldsymbol{z})=\sum^g_{i=1}\sum^n_{j = 1}z_{ij}\Big\{-p\log(2\pi) - \frac{1}{2}\log|\bb{\Sigma}_i|\\ - \frac{w_j}{2}\Big[d_i(\bb{y_j}) + (\bb{u}_j - \bb{c}_i(\bb{y}_j))^T\bb{\Lambda}_i^{-1}(\bb{u}_j - \bb{c}_i(\bb{y}_j))\Big]\Big\}
    \label{(15)}
\end{multline}

\noindent
The parameters seen in Eqs.~\eqref{(13)}-\eqref{(15)}. are derived using the expressions seen below:

\begin{align*}
    &d_i(\bb{y}_j) = (\bb{y}_j - \bb{\mu}_i)^T\Omega_i^{-1}(\bb{y}_j - \bb{\mu}_i)\\
    &\bb{c}_i(\bb{y}_j) = \bb{\Delta}_i^T\Omega_i^{-1}(\bb{y}_j - \bb{\mu}_i)\\
    &\bb{\Lambda}_i = I_p - \bb{\Delta}_i^T\Omega_i^{-1}\bb{\Delta}_i\\
    &\bb{\Omega}_i = \bb{\Sigma}_i + \bb{\Delta}_i\bb{\Delta}_i^T\\
\end{align*}

\vspace{-0.5cm}

\noindent
In the mentioned equations above, $d_i(\bb{y}_j)$ is the squared Mahalanobis distance between $\bb{y}_j$ and $\bb{\mu}_i$ with respect to $\bb{\Omega}_i$ \cite{55}.

\section{Conditional Expected Values of Truncated $t$-Distribution} \label{app_b}
In the E-step of the ERM algorithm, the conditional expected value of two random entities should be calculated, which are $\bb{e}_3$ and $\bb{e}_4$ as seen in Eq.~\eqref{(19)} and Eq.~\eqref{(20)}. These expressions can be calculated as the product of $e_1$ and the first and second moments of the truncated $t$ distribution of vector $\bb{U}$ respectively. For brevity, we denote the defining parameters of the truncated $t$ distribution as $\bb{\mu}$, $\bb{\Sigma}$, $\mathbb{A}$, and $\nu$ as seen in Eq.~\eqref{(26)}.

\begin{equation}
    \bb{U} \sim tt_{p,\nu}(\bb{\mu},\bb{\Sigma};\mathbb{A})
    \label{(26)}
\end{equation}
where the truncated hyperplane is $\mathbb{A} = \{\bb{x} > \bb{a}, \bb{a} \in \mathbb{R}^p\}$. The truncation used in this PDF is left-truncation, meaning that each element of the vector $\bb{x}$ is greater than the associated element in the truncation vector $\bb{a}$. The density of this left truncated distribution is given by Eq.~\eqref{(27)}.

\begin{equation}
    f_{\mathbb{A}}(\bb{x}|\bb{\mu},\bb{\Sigma},\nu) = T^{-1}_{p,\nu}(\bb{a}|\bb{\mu},\bb{\Sigma})t_{p,\nu}(\bb{x}|\bb{\mu},\bb{\Sigma}), \bb{x} \in \mathbb{A}
    \label{(27)}
\end{equation}

For evaluation of the first and second moments of this distribution, O’Hagan \cite{38} first provided the explicit representation of the formulae based on moments of untruncated $t$-distribution in the univariate case of $tt_{p,\nu}(0,\sigma^2;a)$. Later O’Hagan \cite{39} analyzed the multivariate case in the centralized case. Lee et al. \cite{16} then in 2014 generalized the representation of these moments to non-central cases. Table~\ref{tab_6} shows the characteristics of certain notations used in the series of mathematical steps for computing the first and second moment of the truncated $t$-distribution. 

For clarity of the equations mentioned subsequently, a brief description of mathematical notations used in the equations is provided in Table~\ref{tab_6}. In this table, $\bb{x}$ is a $p \times 1$ vector and $\bb{X}$ is a $p \times p$ matrix.

\begin{table}[h]
\renewcommand{\arraystretch}{1.5}
    \centering
        \caption{Description of the notations} \vspace{-0.2cm}
    \begin{tabularx}{\columnwidth}{|c|c|c|>{\raggedright\arraybackslash}X|}
    \hline
         Notation & Type & Size & Description \\ \hline \hline
         $x_i$ & Scalar & $1\times1$ & $i$th element of the vector $\bb{x}$ \\ \hline
         $\bb{x}_{-i}$ & Vector & $(p-1)\times 1$ & $\bb{x}$ without its $i$th element \\ \hline
         $\bb{x}_{-ij}$ & Vector & $(p-2)\times 1$ & $\bb{x}$ without its $i$th and $j$th element \\ \hline
         $\bb{X}_{ij}$ & Matrix & $2\times 2$ & $\bb{X}_{ij} = \begin{bmatrix} x_{ii} & x_{ij} \\ x_{ji} & x_{jj} \end{bmatrix}$ \\ \hline
         $\bb{X}_{-i}$ & Matrix & $(p-1)\times (p-1)$ & Matrix $\bb{X}$ without its $i$th row and column \\ \hline
         $\bb{X}_{-ij}$ & Matrix & $(p-2)\times (p-2)$ & Matrix $\bb{X}$ without its $i$th and $j$th rows and columns \\ \hline
         $\bb{X}_{(i)}$ & Vector & $(p-1)\times 1$ & Vector consisting $i$th column of matrix $\bb{X}$ without its $i$th element \\ \hline
         $\bb{X}_{(ij)}$ & Matrix & $(p-2)\times 2$ & Matrix consisting $i$th and $j$th  columns of matrix $\bb{X}$ without its $i$th and $j$th rows \\ \hline
         
    \end{tabularx}

    \label{tab_6}
\end{table}

\noindent
$E(\bb{X})$ can be represented as the first moment of PDF in Eq.~\eqref{(28)} with:

\begin{equation}
    E(\bb{X}) = \bb{\mu} + \bb{\epsilon}
    \label{(28)}
\end{equation}

\begin{equation}
    \bb{\epsilon} = c_1^{-1}\bb{\Sigma}\bb{\xi}
\end{equation}

\begin{equation}
    c_1 = T_{p,\nu}(\bb{\mu - a}|\bb{0},\bb{\Sigma})
\end{equation}

\begin{multline}
    \xi_i = (2\pi\sigma_{ii})^{-\frac{1}{2}}\big(\frac{\nu}{\nu + \sigma_{ii}^{-1}(\mu_i - a_i)^2}\big)^{(\frac{\nu - 1}{2})}\\ \sqrt{\frac{\nu}{2}}\frac{\Gamma(\frac{\nu - 1}{2})}{\Gamma (\frac{\nu}{2})} T_{p-1,\nu-1}(\bb{a}^{\star}|\bb{\mu},\bb{\Sigma}^{\star})
\end{multline}
\noindent
where 

\begin{equation}
    \bb{a}^{\star} = (\bb{\mu}_{-i} - \bb{a}_{-i}) - (\mu_i - a_i)\sigma_{ii}^{-1}\bb{\Sigma}_{(i)}
\end{equation}

\begin{equation}
    \bb{\Sigma}^{\star} = \Big(\frac{\nu + \sigma_{ii}^{-1}(\mu_i - a_i)^2}{\nu - 1}\Big)(\bb{\Sigma}_{-i} - \sigma_{ii}^{-1}\bb{\Sigma}_{(i)}\bb{\Sigma}_{(i)}^T)
\end{equation}

However, For the second moment, the equations become more complex as seen in Eq.~\eqref{(34)}.

\begin{multline}
    E(\bb{XX}^T) = \bb{\mu\mu}^T + \bb{\mu\epsilon}^T + \bb{\epsilon\mu}^T\\ - c_1^{-1}\bb{\Sigma H \Sigma} + c_2 c_1^{-1} \big(\frac{\nu}{\nu - 2}\big) \bb{\Sigma}
    \label{(34)}
\end{multline}

\begin{equation}
    c_2 = T_{p,\nu-2}\big(\bb{\mu - a}|\bb{0}, \bb{\Sigma}\big)
\end{equation}
\noindent
where $H$ is a $p\times p$ matrix for which we first calculate the off-diagonal elements as the following:

\begin{multline}
    h_{ij} = \frac{1}{2\pi\sqrt{\sigma_{ii}\sigma_{jj} - \sigma_{ij}^2}}\big(\frac{\nu}{\nu - 2}\big)\big(\frac{\nu}{\nu^{\star}}\big)^{(\frac{\nu - 2}{2})}\\
    T_{p-2,\nu-2}(\bb{a}^{\star\star}|\bb{0},\bb{\Sigma}^{\star\star}), i \neq j
\end{multline}
\noindent
and the diagonal elements are calculated as such:

\begin{equation}
    h_{ii} = \sigma^{-1}_ii \bigg((\mu_i - a_i)\xi_i - \sum_{i\neq j}\sigma_{ij}h_{ij}\bigg)
\end{equation}
\noindent
where

\begin{equation}
    \nu^{\star} = \nu + (\bb{\mu}_{ij} - \bb{a}_{ij})^T\bb{\Sigma}_{ij}^{-1}(\bb{\mu}_{ij} - \bb{a}_{ij})
\end{equation}

\begin{equation}
    \bb{a}^{\star\star} = (\bb{\mu}_{-ij} - \bb{a}_{-ij}) - \bb{\Sigma}_{(ij)}\bb{\Sigma}_{ij}^{-1}(\bb{\mu}_{ij} - \bb{a}_{ij})
\end{equation}

\begin{equation}
    \bb{\Sigma}^{\star\star} = \frac{\nu^{\star}}{\nu - 2}(\bb{\Sigma}_{-ij} - \bb{\Sigma}_{(ij)}\bb{\Sigma}_{ij}^{-1}\bb{\Sigma}_{(ij)}^T)
\end{equation}
\noindent
It is important to mention that in the case of univariate datasets, $T_{p-1,v-1}(\bb{a}^{\star}|\bb{0},\bb{\Sigma}^{\star})=1$, and in the case of univariate and bivariate datasets, $T_{p-2,v-2}(\bb{a}^{\star\star}|\bb{0},\bb{\Sigma}^{\star\star})=1$.

% Can use something like this to put references on a page
% by themselves when using endfloat and the captionsoff option.
\ifCLASSOPTIONcaptionsoff
  \newpage
\fi

\bibliographystyle{IEEEtran}
\bibliography{ref}

% Generated by IEEEtran.bst, version: 1.14 (2015/08/26)
\begin{thebibliography}{10}
\providecommand{\url}[1]{#1}
\csname url@samestyle\endcsname
\providecommand{\newblock}{\relax}
\providecommand{\bibinfo}[2]{#2}
\providecommand{\BIBentrySTDinterwordspacing}{\spaceskip=0pt\relax}
\providecommand{\BIBentryALTinterwordstretchfactor}{4}
\providecommand{\BIBentryALTinterwordspacing}{\spaceskip=\fontdimen2\font plus
\BIBentryALTinterwordstretchfactor\fontdimen3\font minus
  \fontdimen4\font\relax}
\providecommand{\BIBforeignlanguage}[2]{{%
\expandafter\ifx\csname l@#1\endcsname\relax
\typeout{** WARNING: IEEEtran.bst: No hyphenation pattern has been}%
\typeout{** loaded for the language `#1'. Using the pattern for}%
\typeout{** the default language instead.}%
\else
\language=\csname l@#1\endcsname
\fi
#2}}
\providecommand{\BIBdecl}{\relax}
\BIBdecl

\bibitem{61}
Z.~Liu, L.~Yu, J.~H. Hsiao, and A.~B. Chan, ``Primal-gmm: Parametric manifold
  learning of gaussian mixture models,'' \emph{IEEE Transactions on Pattern
  Analysis and Machine Intelligence}, vol.~44, no.~6, pp. 3197--3211, 2021.

\bibitem{62}
R.~P. Browne, P.~D. McNicholas, and M.~D. Sparling, ``Model-based learning
  using a mixture of mixtures of gaussian and uniform distributions,''
  \emph{IEEE Transactions on Pattern Analysis and Machine Intelligence},
  vol.~34, no.~4, pp. 814--817, 2011.

\bibitem{63}
L.~Yu, T.~Yang, and A.~B. Chan, ``Density-preserving hierarchical em algorithm:
  Simplifying gaussian mixture models for approximate inference,'' \emph{IEEE
  transactions on pattern analysis and machine intelligence}, vol.~41, no.~6,
  pp. 1323--1337, 2018.

\bibitem{65}
W.~Fan, L.~Yang, and N.~Bouguila, ``Unsupervised grouped axial data modeling
  via hierarchical bayesian nonparametric models with watson distributions,''
  \emph{IEEE Transactions on Pattern Analysis and Machine Intelligence},
  vol.~44, no.~12, pp. 9654--9668, 2021.

\bibitem{4}
M.~Karami and L.~Wang, ``Fault detection and diagnosis for nonlinear systems: A
  new adaptive gaussian mixture modeling approach,'' \emph{Energy and
  Buildings}, vol. 166, pp. 477--488, 2018.

\bibitem{52}
H.-C. Yan, J.-H. Zhou, and C.~K. Pang, ``Gaussian mixture model using
  semisupervised learning for probabilistic fault diagnosis under new data
  categories,'' \emph{IEEE Transactions on Instrumentation and Measurement},
  vol.~66, no.~4, pp. 723--733, 2017.

\bibitem{53}
Y.~Hong, M.~Kim, H.~Lee, J.~J. Park, and D.~Lee, ``Early fault diagnosis and
  classification of ball bearing using enhanced kurtogram and gaussian mixture
  model,'' \emph{IEEE Transactions on Instrumentation and Measurement},
  vol.~68, no.~12, pp. 4746--4755, 2019.

\bibitem{2}
A.~Das, U.~R. Acharya, S.~S. Panda, and S.~Sabut, ``Deep learning based liver
  cancer detection using watershed transform and gaussian mixture model
  techniques,'' \emph{Cognitive Systems Research}, vol.~54, pp. 165--175, 2019.

\bibitem{3}
A.~M. Khan, H.~El-Daly, and N.~M. Rajpoot, ``A gamma-gaussian mixture model for
  detection of mitotic cells in breast cancer histopathology images,'' in
  \emph{Proceedings of the 21st International Conference on Pattern Recognition
  (ICPR2012)}.\hskip 1em plus 0.5em minus 0.4em\relax IEEE, 2012, pp. 149--152.

\bibitem{1}
M.~Najafi, M.~Sharifi, K.~Adams, and M.~Tavakoli, ``Robotic assistance for
  children with cerebral palsy based on learning from tele-cooperative
  demonstration,'' \emph{International Journal of Intelligent Robotics and
  Applications}, vol.~1, no.~1, pp. 43--54, 2017.

\bibitem{40}
D.~A. Duque, F.~A. Prieto, and J.~G. Hoyos, ``Trajectory generation for robotic
  assembly operations using learning by demonstration,'' \emph{Robotics and
  Computer-Integrated Manufacturing}, vol.~57, pp. 292--302, 2019.

\bibitem{58}
L.~Rozo~Casta{\~n}eda, P.~Jimenez~Schlegl, and C.~Torras, ``Sharpening haptic
  inputs for teaching a manipulation skill to a robot,'' in \emph{1st IEEE
  International Conference on Applied Bionics and Biomechanics}, 2010, pp.
  331--340.

\bibitem{59}
L.~D. Rozo, P.~Jim{\'e}nez, and C.~Torras, ``Learning force-based robot skills
  from haptic demonstration.'' in \emph{CCIA}, 2010, pp. 331--340.

\bibitem{6}
J.~A. Hartigan, \emph{Clustering algorithms}.\hskip 1em plus 0.5em minus
  0.4em\relax John Wiley \& Sons, Inc., 1975.

\bibitem{7}
R.~C. Rose and D.~A. Reynolds, ``Text independent speaker identification using
  automatic acoustic segmentation,'' in \emph{International Conference on
  Acoustics, Speech, and Signal Processing}.\hskip 1em plus 0.5em minus
  0.4em\relax IEEE, 1990, pp. 293--296.

\bibitem{8}
D.~A. Reynolds and R.~C. Rose, ``Robust text-independent speaker identification
  using gaussian mixture speaker models,'' \emph{IEEE transactions on speech
  and audio processing}, vol.~3, no.~1, pp. 72--83, 1995.

\bibitem{9}
D.~A. Reynolds, T.~F. Quatieri, and R.~B. Dunn, ``Speaker verification using
  adapted gaussian mixture models,'' \emph{Digital signal processing}, vol.~10,
  no. 1-3, pp. 19--41, 2000.

\bibitem{10}
A.~M. Alqudah, ``An enhanced method for real-time modelling of cardiac related
  biosignals using gaussian mixtures,'' \emph{Journal of medical engineering \&
  technology}, vol.~41, no.~8, pp. 600--611, 2017.

\bibitem{11}
A.~Pervez, A.~Ali, J.-H. Ryu, and D.~Lee, ``Novel learning from demonstration
  approach for repetitive teleoperation tasks,'' in \emph{2017 IEEE World
  Haptics Conference (WHC)}.\hskip 1em plus 0.5em minus 0.4em\relax IEEE, 2017,
  pp. 60--65.

\bibitem{41}
B.~Akgun, M.~Cakmak, K.~Jiang, and A.~L. Thomaz, ``Keyframe-based learning from
  demonstration,'' \emph{International Journal of Social Robotics}, vol.~4,
  no.~4, pp. 343--355, 2012.

\bibitem{12}
Y.~Li, Z.~Wang, T.~Zhao, and S.~Wanqing, ``Research on a pattern recognition
  method of cyclic gmm-fcm based on joint time-domain features,'' \emph{IEEE
  Access}, vol.~9, pp. 1904--1917, 2020.

\bibitem{42}
G.~Chen and S.~Luo, ``Robust bayesian hierarchical model using
  normal/independent distributions,'' \emph{Biometrical Journal}, vol.~58,
  no.~4, pp. 831--851, 2016.

\bibitem{13}
A.~P. Dempster, N.~M. Laird, and D.~B. Rubin, ``Maximum likelihood from
  incomplete data via the em algorithm,'' \emph{Journal of the Royal
  Statistical Society: Series B (Methodological)}, vol.~39, no.~1, pp. 1--22,
  1977.

\bibitem{26}
S.~K. Sahu, D.~K. Dey, and M.~D. Branco, ``A new class of multivariate skew
  distributions with applications to bayesian regression models,''
  \emph{Canadian Journal of Statistics}, vol.~31, no.~2, pp. 129--150, 2003.

\bibitem{23}
R.~P. Browne and P.~D. McNicholas, ``A mixture of generalized hyperbolic
  distributions,'' \emph{Canadian Journal of Statistics}, vol.~43, no.~2, pp.
  176--198, 2015.

\bibitem{24}
P.~Spurek, ``General split gaussian cross--entropy clustering,'' \emph{Expert
  Systems with Applications}, vol.~68, pp. 58--68, 2017.

\bibitem{64}
Y.~Wei, Y.~Tang, and P.~D. McNicholas, ``Flexible high-dimensional unsupervised
  learning with missing data,'' \emph{IEEE transactions on pattern analysis and
  machine intelligence}, vol.~42, no.~3, pp. 610--621, 2018.

\bibitem{25}
G.~J. McLachlan, S.~X. Lee, and S.~I. Rathnayake, ``Finite mixture models,''
  \emph{Annual review of statistics and its application}, vol.~6, pp. 355--378,
  2019.

\bibitem{14}
T.~I. Lin, ``Maximum likelihood estimation for multivariate skew normal mixture
  models,'' \emph{Journal of Multivariate Analysis}, vol. 100, no.~2, pp.
  257--265, 2009.

\bibitem{16}
S.~Lee and G.~J. McLachlan, ``Finite mixtures of multivariate skew
  t-distributions: some recent and new results,'' \emph{Statistics and
  Computing}, vol.~24, no.~2, pp. 181--202, 2014.

\bibitem{54}
G.~J. McLachlan and D.~Peel, ``Robust cluster analysis via mixtures of
  multivariate t-distributions,'' in \emph{Joint IAPR International Workshops
  on Statistical Techniques in Pattern Recognition (SPR) and Structural and
  Syntactic Pattern Recognition (SSPR)}.\hskip 1em plus 0.5em minus 0.4em\relax
  Springer, 1998, pp. 658--666.

\bibitem{51}
A.~Azzalini and A.~Capitanio, ``Distributions generated by perturbation of
  symmetry with emphasis on a multivariate skew t-distribution,'' \emph{Journal
  of the Royal Statistical Society: Series B (Statistical Methodology)},
  vol.~65, no.~2, pp. 367--389, 2003.

\bibitem{21}
S.~Pyne, X.~Hu, K.~Wang, E.~Rossin, T.-I. Lin, L.~M. Maier, C.~Baecher-Allan,
  G.~J. McLachlan, P.~Tamayo, D.~A. Hafler \emph{et~al.}, ``Automated
  high-dimensional flow cytometric data analysis,'' \emph{Proceedings of the
  National Academy of Sciences}, vol. 106, no.~21, pp. 8519--8524, 2009.

\bibitem{18}
C.~R.~B. Cabral, V.~H. Lachos, and M.~O. Prates, ``Multivariate mixture
  modeling using skew-normal independent distributions,'' \emph{Computational
  Statistics \& Data Analysis}, vol.~56, no.~1, pp. 126--142, 2012.

\bibitem{20}
I.~Vrbik and P.~McNicholas, ``Analytic calculations for the em algorithm for
  multivariate skew-t mixture models,'' \emph{Statistics \& Probability
  Letters}, vol.~82, no.~6, pp. 1169--1174, 2012.

\bibitem{15}
T.-I. Lin, ``Robust mixture modeling using multivariate skew t distributions,''
  \emph{Statistics and Computing}, vol.~20, no.~3, pp. 343--356, 2010.

\bibitem{43}
G.~C. Wei and M.~A. Tanner, ``A monte carlo implementation of the em algorithm
  and the poor man's data augmentation algorithms,'' \emph{Journal of the
  American statistical Association}, vol.~85, no. 411, pp. 699--704, 1990.

\bibitem{19}
S.~Lee and G.~J. McLachlan, ``On the fitting of mixtures of multivariate skew
  t-distributions via the em algorithm,'' \emph{arXiv preprint
  arXiv:1109.4706}, 2011.

\bibitem{49}
P.~J. Green, ``On use of the em algorithm for penalized likelihood
  estimation,'' \emph{Journal of the Royal Statistical Society: Series B
  (Methodological)}, vol.~52, no.~3, pp. 443--452, 1990.

\bibitem{48}
H.~J. Ho, T.-I. Lin, H.-Y. Chen, and W.-L. Wang, ``Some results on the
  truncated multivariate t distribution,'' \emph{Journal of Statistical
  Planning and Inference}, vol. 142, no.~1, pp. 25--40, 2012.

\bibitem{22}
S.~X. Lee and G.~J. McLachlan, ``Finite mixures of canonical fundamenal skew
  \$\$\$\$-disribuions,'' \emph{Statistics and computing}, vol.~26, no.~3, pp.
  573--589, 2016.

\bibitem{57}
G.~J. McLachlan and T.~Krishnan, \emph{The EM algorithm and extensions}.\hskip
  1em plus 0.5em minus 0.4em\relax John Wiley \& Sons, 2007.

\bibitem{29}
A.~Genz and F.~Bretz, ``Comparison of methods for the computation of
  multivariate t probabilities,'' \emph{Journal of Computational and Graphical
  Statistics}, vol.~11, no.~4, pp. 950--971, 2002.

\bibitem{35}
S.~X. Lee and G.~J. McLachlan, ``Emmix-uskew: an r package for fitting mixtures
  of multivariate skew t-distributions via the em algorithm,'' \emph{arXiv
  preprint arXiv:1211.5290}, 2012.

\bibitem{47}
R.~Levy, ``Probabilistic models in the study of language,'' \emph{Online Draft,
  Nov}, 2012.

\bibitem{31}
P.~J. Green, ``On use of the em algorithm for penalized likelihood
  estimation,'' \emph{Journal of the Royal Statistical Society: Series B
  (Methodological)}, vol.~52, no.~3, pp. 443--452, 1990.

\bibitem{32}
S.~Lee and G.~J. McLachlan, ``On the fitting of mixtures of multivariate skew
  t-distributions via the em algorithm,'' \emph{arXiv preprint
  arXiv:1109.4706}, 2011.

\bibitem{33}
H.~J. Ho, T.-I. Lin, H.-Y. Chen, and W.-L. Wang, ``Some results on the
  truncated multivariate t distribution,'' \emph{Journal of Statistical
  Planning and Inference}, vol. 142, no.~1, pp. 25--40, 2012.

\bibitem{36}
R.~D. Cook and S.~Weisberg, \emph{An introduction to regression
  graphics}.\hskip 1em plus 0.5em minus 0.4em\relax John Wiley \& Sons, 2009,
  vol. 405.

\bibitem{56}
R.~O’Keeffe, S.~Y. Shirazi, S.~Mehrdad, T.~Crosby, A.~M. Johnson, and S.~F.
  Atashzar, ``Perilaryngeal-cranial functional muscle network differentiates
  vocal tasks: A multi-channel semg approach,'' \emph{IEEE Transactions on
  Biomedical Engineering}, vol.~69, no.~12, pp. 3678--3688, 2022.

\bibitem{60}
R.~O'Keeffe, Y.~Shirazi, S.~Mehrdad, T.~Crosby, A.~M. Johnson, and S.~F.
  Atashzar, ``Perilaryngeal functional muscle network in patients with vocal
  hyperfunction-a case study,'' \emph{bioRxiv}, pp. 2023--01, 2023.

\bibitem{55}
A.~Basharat, A.~Gritai, and M.~Shah, ``Learning object motion patterns for
  anomaly detection and improved object detection,'' in \emph{2008 IEEE
  Conference on Computer Vision and Pattern Recognition}.\hskip 1em plus 0.5em
  minus 0.4em\relax IEEE, 2008, pp. 1--8.

\bibitem{38}
A.~O'hagan, ``Bayes estimation of a convex quadratic,'' \emph{Biometrika},
  vol.~60, no.~3, pp. 565--571, 1973.

\bibitem{39}
A.~O’Hagan, ``Moments of the truncated multivariatet distribution,'' 1976.

\end{thebibliography}
\end{document}